\documentclass[conference]{IEEEtran}
\usepackage{cite}
\usepackage{amsmath,amssymb,amsfonts,amsthm}
\usepackage{algorithm}
\usepackage{algorithmic}
\usepackage{graphicx}
\usepackage{textcomp}
\usepackage{enumitem}
\usepackage{nicefrac}
\usepackage{multirow}
\usepackage{tabularx}
\usepackage{soul}
\usepackage{xcolor}
\usepackage{subfigure}
\usepackage{url}
\usepackage{bm}

\newcolumntype{C}[1]{>{\centering\arraybackslash}p{#1}}

\def\BibTeX{{\rm B\kern-.05em{\sc i\kern-.025em b}\kern-.08em
		T\kern-.1667em\lower.7ex\hbox{E}\kern-.125emX}}

\newtheorem{prop}{Proposition}
\newtheorem{definition}{Definition}
\newtheorem{remark}{Remark}[section]

\usepackage{booktabs}
\newcommand{\diag}[1]{ \text{diag}\!\left(#1\right)}

\title{AutoSF:
	Searching Scoring Functions for  
	Knowledge Graph Embedding}

\author{\IEEEauthorblockN{Yongqi Zhang\IEEEauthorrefmark{2},
			Quanming Yao\IEEEauthorrefmark{3}, Wenyuan Dai\IEEEauthorrefmark{3} and
			Lei Chen\IEEEauthorrefmark{2}}
		\IEEEauthorblockA{\IEEEauthorrefmark{2}The Hong Kong University of Science and Technology, Hong Kong SAR, China\\
			\IEEEauthorrefmark{3}4Paradigm Inc., Beijing, China\\
			\IEEEauthorrefmark{2}\{yzhangee,leichen\}@cse.ust.hk,
			\IEEEauthorrefmark{3}\{yaoquanming,daiwenyuan\}@4paradigm.com}}
\begin{document}	

\maketitle

\begin{abstract}
Scoring functions (SFs), which measure the plausibility
of triplets in knowledge graph (KG), have become the crux of KG embedding. 
Lots of SFs,
which target at capturing different kinds of relations in KGs,
have been designed by humans in recent years.
However, 
as relations can exhibit complex patterns that are hard to infer before training,
none of them can consistently perform better than others on existing benchmark data sets. 
In this paper,
inspired by the recent success of automated machine learning (AutoML), 
%instead of proposing another new SF,
we propose to automatically design SFs (AutoSF) for distinct KGs by the AutoML techniques.
However,
it is non-trivial to explore domain-specific information here to make AutoSF efficient and effective.
We firstly identify a unified representation over popularly used SFs, 
which helps to set up a search space for AutoSF. 
Then, we propose a greedy algorithm to search in such a space efficiently.
The algorithm
is further sped up by a filter and a predictor,
which can avoid repeatedly training SFs with same expressive ability 
and
help removing bad candidates during the search before model training.
Finally,
we perform extensive experiments on benchmark data sets.
Results on link prediction and triplets classification
show that the searched SFs by AutoSF,
are KG dependent, new to the literature, and outperform the state-of-the-art SFs designed by humans.
\end{abstract}

\setcounter{figure}{0}
\setcounter{table}{0}

\section{Introduction}
\label{sec:intro}
%%\vspace{-5px}

Knowledge Graph (KG) \cite{singhal2012introducing,nickel2016review,wang2017knowledge}, 
as a special kind of graph structure with entities as nodes and relations as edges, 
is important to both data mining and machine learning,
and has inspired various downstream applications, 
e.g., structured search \cite{singhal2012introducing,dong2014knowledge,tang2016aminer},
question answering \cite{lukovnikov2017neural} and
recommendation \cite{zhang2016collaborative}.
%and few-shot learning \cite{wang2018zero}. 
%to indicate that two entities $h$ (head) and $t$ (tail) are connected by a relation $r$,
%e.g. \textit{(NewYork, isLocatedIn, USA)}
%\cite{singhal2012introducing}.
%A number of large scale KGs are established in last decades, 
%such as WordNet \cite{miller1995wordnet}, Freebase \cite{bollacker2008freebase}, 
%DBpedia \cite{auer2007dbpedia}, YAGO \cite{suchanek2007yago}, etc.
%They have improved 
In KGs, each edge is represented as a triplet with form \textit{(head entity, relation, tail entity)},
denoted as $(h,r,t)$.
A fundamental issue is how to quantize
the plausibility 
%\footnote{$\surd$ + change all similarity into plausibility}
of triplets $(h, r, t)$s
%such that subsequent applications can be performed
\cite{getoor2007introduction,wang2017knowledge}.
KG embedding (KGE) has recently emerged and been developed as 
a promising method serving this purpose
%\footnote{$\surd$ it is OK. + remove some of the ref, too many}
\cite{nickel2011three,yang2014embedding,nickel2016holographic,dettmers2017convolutional,trouillon2017knowledge,liu2017analogical,kazemi2018simple,lacroix2018canonical,zhang2018nscaching}. 
Basically,
given a set of observed triplets, 
KGE attempts to learn low-dimensional vector representations of entities and relations
so that the plausibility of triplets can be quantized.
Scoring function (SF),
which returns a score for $(h,r,t)$ based on the embeddings,
is used to measure the plausibility.
%The plausibility is measured by a scoring function (SF),
%which returns a score for $(h,r,t)$ based on the embeddings.
Generally,
SF is designed and chosen by humans
and it has significant effects on embeddings' quality \cite{nickel2016review,wang2017knowledge,lin2018knowledge}.
%and is also the most important perspective of KGE,
%which can significantly affect embeddings' quality \cite{nickel2016review,wang2017knowledge,lin2018knowledge}.

%\footnote{$\surd$ the logic in this paragraph is not tight.
%	you can re-write it based on Sec.\ref{ssec:score}.
%	say we have three directions;
%	BLM is the most promising one.}
Ever since the invention of KGE, 
many SFs have been proposed in the literature.
Let the embedded vectors of
$h$,
$r$ and $t$ 
be $\bm{h}$,
$\bm{r}$ and
$\bm{t}$ respectively.
%Typical examples
%include 
TransE \cite{bordes2013translating},
a representative embedding model,
interprets the triplet $(h,r,t)$ as a translation $r$
from head entity $h$ to tail entity $t$,
i.e.  
the embeddings satisfy $\bm{h} + \bm{r} = \bm{t}$.
%\footnote{$\surd$ what problems these variants have solved?}
Variants of TransE like 
TransH \cite{wang2014knowledge} and
TransR \cite{lin2015learning},
project the embedding vector into different spaces and 
enables the embedding to model relationships that are one-to-many, many-to-one or many-to-many.
These models are categorized into translational distance models (TDMs).
However, as proved in \cite{wang2017multi,wang2018evaluating},
TDMs are not fully expressive 
and their empirical performance is inferior to other models.
%\footnote{$\surd$ need to add 1-2 sentences here}
%TransE \cite{bordes2013translating}
%and its variants like
%%such as
%TransH \cite{wang2014knowledge} and
%TransR \cite{lin2015learning},
%which belong to translational distance models (TDMs),
%model the plausibility by using projected distance in vector space.
RESCAL \cite{nickel2011three},
DistMult \cite{yang2014embedding},
ComplEx \cite{trouillon2017knowledge},
Analogy \cite{liu2017analogical}
and more recently proposed SimplE \cite{kazemi2018simple,lacroix2018canonical},
%belongs to bilinear models (BLMs).
%BLMs 
use a bilinear function $\bm{h}^\top\bm{R}\;\bm{t}$ to model the plausibility of triplets,
where $\bm{R}$ is a square matrix related to relation embedding.
These models belong to the bilinear model (BLMs).
Different BLMs use different constrains to regularize the relation matrix $\bm R$
in order to adapt to different datasets.
Inspired by the success of deep networks \cite{bengio2013representation},
some neural network models (NNMs) have also been explored as SFs,
e.g., MLP \cite{dong2014knowledge}, 
NTM \cite{socher2013reasoning},
Neural LP \cite{yang2017differentiable}
and ConvE \cite{dettmers2017convolutional}.
Even though neural networks are powerful and have strong expressive ability,
the NNMs do not perform well in KGE domain because of not being well-regularized.

Among the existing SFs,
BLM-based ones
%e.g. CompEx 
%\cite{trouillon2017knowledge} 
%and SimplE,
%\cite{lacroix2018canonical,kazemi2018simple}, 
are the most powerful 
as indicated by both the state-of-the-art results \cite{lacroix2018canonical} and theoretical guarantees on expressiveness \cite{wang2017multi, kazemi2018simple}.
However,
since
different KGs have distinct patterns in relations \cite{paulheim2017knowledge},
a SF which adapts well to one KG may not perform consistently well on other KGs.
Besides,
designing new SFs to outperform state-of-the-art SFs 
is challenging.
Therefore,
%with so many SFs,
how to choose and design a good SF for a certain KG is a non-trivial and difficult problem.

%\footnote{$\surd$ need to expand this section, make it longer.}
Recently,
automated machine learning (AutoML) \cite{quanming2018auto,automl_book}
has exhibited its power in
many machine learning tasks and applications,
e.g. model selection \cite{feurer2015efficient},
image classification \cite{liu2018darts,yao2019differentiable}
and recommendation \cite{yao2019efficient}.
In order to select proper models and hyper-parameters for different tasks,
hyperparameter optimization (HPO) has been proposed \cite{feurer2015efficient,falkner2018bohb}
to effectively and efficiently find better configurations,
which previously requires great human efforts.
Another hot trend in AutoML is to search better neural networks for deep learning models.
Neural architecture search (NAS) \cite{zoph2017neural} 
has identified networks with fewer parameters and better performance than networks designed by humans.

Inspired by the success of AutoML,
we aim to design better and novel data-dependent SFs for KGE.
Specifically,
we propose automated scoring function (AutoSF)
which can automatically search an SF for a given KG.
It can not only reduce human's effort in designing new SFs,
but also make adaptation to different KGs.
However,
it is not easy to achieve the above goal.
When applying AutoML,
two important perspectives,
i.e. search space, which helps to figure out important properties of the underling problems,
and search algorithm, which determines the efficiency of finding better points in the space,
need serious consideration.
In this work,
%with the consideration of search space and search algorithm
we have made the following contributions to achieve the goals:
\begin{itemize}[leftmargin = 10px,itemsep=1px]
\item 
First,
we make an important observation over existing SFs,
which allows us to represent the BLM-based SFs in a unified form.
%among existing SFs in BLMs,
%which are more powerful than TDMs and NNMs.
%and represent
Based on the unified representation,
we formulate the design of SFs as an AutoML problem (i.e. AutoSF),
and set up the corresponding search space.
The space is not only specific enough
to cover good SFs designed by humans,
but also general enough to include 
novel SFs not visited in the literature.

\item 
Second, we observe it is common that different KGs have distinct properties
on relations
that are
symmetric, asymmetric, inverse, etc.
This inspires us to conduct domain-specific analysis on the KGE models,
and design constraints 
%	on their expressiveness
to effectively guide subsequent searches in the space.

\item 
Third, 
we propose a progressive greedy algorithm to search through such a space.
We further build a filter to avoid training redundant SFs
and a predictor with specifically designed symmetry-related features (SRF) to select promising SFs.
The search algorithm can significantly reduce the search space size
by capturing the domain specific properties of candidate SFs.

\item
Finally, experimental results on five popular benchmarks on link prediction and triplet classification tasks
demonstrate that the SFs searched by AutoSF
%Finally, we conduct experiments on five popular benchmarks on link prediction and triplet classification tasks. 
%Experimental results demonstrate that SFs searched by AutoSF
outperform the start-of-the-art SFs
designed by humans.
In addition,
the searched SFs
are KG dependent and new to the literature.
We further conduct case study on the searched SFs 
to provide means for analyzing KGs,
which can inspire better understanding of embedding techniques for future researches.
%\footnote{$\surd$
%	mention that we have also done case study in Sec.\ref{ssec:casestufy},
%	which can help designing of SFs in the future.
%	Also check Remark~\ref{rmk:reg}.}
\end{itemize}

{
\noindent
\textit{Notations.}
We denote vectors by lowercase boldface, and matrix by uppercase boldface.
A KG contains a set of triplets $\mathcal S=\{(h, r, t)\}$ with $h,t\in\mathcal E$ and $r\in\mathcal R$,
where $\mathcal E$ and $\mathcal R$ are the set of entities and relations, respectively.
%the most frequently used symbols are summarized in Tab.~\ref{tab:notation}.
%For a KG, the set of entities and relations are given by $\mathcal E$ and $\mathcal R$, respectively.
%The set of negative triplets for each triplet $(h, r, t)$ is denoted as $\tilde{S}_{(h,r,t)}=\{(\tilde{h}, r, \tilde{t}) \}$.
%The trainable embeddings for entities are $\bm e\in\mathbb R^d$ and relations are $\bm r \in \mathbb R^d$.
For simplicity, the embeddings are represented by letters of indices in boldface, 
e.g. $\bm h$, $\bm r$, $\bm t$ are embeddings of $h, r, t$, respectively,
and $\bm h, \bm t$ share the same set of embedding parameters $\bm e$.
%When computing the scores,
$\langle \bm a,\bm b, \bm c\rangle=\sum_{i=1}^d \bm a_i\cdot \bm b_i \cdot \bm c_i$ is the triple dot product 
and can be alternatively represented as $\bm a^\top \diag{\bm b}\bm c$,
where $\diag{\bm b}=\mathbf D^{\bm b}\in\mathbb R^{d\times d}$ is the diagonal matrix of $\bm b$.
We denote the complex vector $\bm v\!=\!\bm v_{re} \!+ \!i\bm v_{im}\in\mathbb C^d$ with $\bm v_{re}, \bm v_{im}\!\in\!\mathbb R^d$.
The conjugate of a complex vector is $conj(\bm v)=\bm v_{re} - i\bm v_{im}$.
%For a real-valued vector $\bm v$,
%$\bm v=[ \bm v_1, \bm v_2, \dots, \bm v_n ]$
%means the split of $d$-dimensional vector $\bm v$ into $n$ $\nicefrac{d}{n}$-dimensional vectors $\bm v_i$.
%In addition,
%we list the frequently used acronyms and their full names in Tab.~\ref{tab:acronym}.
}

\section{Related Works}
\label{sec:relworks}

%For a KG,
%its entity and relation set are given by $\mathcal E$ and $\mathcal R$, respectively. 
%A triplet in the KG is given by $(h, r, t)$,
%where $h,t \in \mathcal E$ are indices of the head and tail entity, 
%and $r \in \mathcal R$ is the index of the relation.
%Embedding parameters of a KGE model are given as $\bm{e}$ for each entity and $\bm{r}$ for each relation.
%For simplicity,
%the embeddings in this paper
%are represented by letters of indices in boldface,
%e.g.,
%$\bm{h}$, $\bm{r}, \bm{t}$ are embeddings of $h, r, t$, respectively,
%and 
%$\bm{h}, \bm{t}$ share the same set of embedding parameters $\bm{e}$.
%%$\bm{h}$, $\bm{r}$, $\bm{t}$ are embedding vectors of $h$, $r$, $t$ respectively.
%$\left\langle \bm{a}, \bm{b}, \bm{c} \right\rangle$
%is the dot product and is equal to $\bm{a}^\top\,\diag{\bm{b}} \bm{c}$ for real-valued vectors,
%and is the \textit{Hermitian} product for complex-valued vectors as in \cite{trouillon2017knowledge}.
%The diagonal matrix $\diag{\bm{b}}$ is constructed with elements in $\bm b$.
%$\text{conj}({\bm{t}})$ is the complex conjugate of $\bm{t}\in \mathbb C^d$.
%Finally,
%$f(\bm{h}, \bm{r}, \bm{t})$ is the scoring function (SF),
%which returns 
%a real value representing
%the plausibility for triplet $(h,r,t)$,
%%between the head entity $h$ and tail entity $t$ when given relation $r$,
%and the higher score indicates the better plausibility.

\subsection{Knowledge graph embedding (KGE)}
\label{ssec:score}

\begin{table*}[ht]
	\centering
	\caption{Existing SFs covered by our search space.
		For Analogy and SimplE,
		the embedding splits into two parts,
		i.e.  $\bm{h}^{\top} \!\! = [ \hat{\bm{h}}^{\top}, \breve{\bm{h}}^{\top} ]$
		and $d \! = \hat{d} + \breve{d}$ (same for $\bm{r}$ and $\bm{t}$).
		The relation types are summarized in Tab.~\ref{tab:comrel}.
	}
	\vspace{-7px}
	\label{tab:scofun}
	\renewcommand{\arraystretch}{1.4}
	%		\begin{tabular}{p{2cm}<{\centering}|p{2cm}<{\centering}|p{3cm}<{\centering}}
	\begin{tabular}{c|c|C{4cm}|c}
		\hline
		scoring function                               &                                                                   embeddings                                                                    &                                                                                                 definition                                            &      relation types that can model                    \\ \hline
		DistMult \cite{yang2014embedding}                      &            symmetric                      $\bm{h}, \bm{r}, \bm{t}\! \in\!\mathbb R^d$                                                   &                                                                             $\left\langle \bm{h}, \bm{r},  \bm{t}\right\rangle$                                                         &      symmetric               \\ \hline
		{ComplEx \cite{trouillon2017knowledge} / HolE \cite{nickel2016holographic}} &                                          $\bm{h}, \bm{r}, \bm{t} \!\in\!\mathbb C^d$                                          &                                                     Re$\left( \left\langle \bm{h},\bm{r}, \text{conj}({\bm{t}})\right\rangle \right)$                                       &    symmetric, anti-symmetric, asymmetric, inverse       \\ \hline
		Analogy \cite{liu2017analogical}              & $ \hat{\bm{h}},\hat{\bm{r}},\hat{\bm{t}}\!\in\!\mathbb R^{\hat{d}}$, $\breve{\bm{h}},\breve{\bm{r}},\breve{\bm{t}}\!\in\!\mathbb C^{\breve{d}}$ & $\!\! \left\langle \hat{\bm{h}}, \hat{\bm{r}}, \hat{\bm{t}}\right\rangle$ + Re$\left( \left\langle  \breve{\bm{h}}, \breve{\bm{r}}, \text{conj}(\breve{\bm{t}})\right\rangle \right) \!\!$ &	symmetric, anti-symmetric, asymmetric, inverse	\\ \hline
		{SimplE \cite{kazemi2018simple} / CP \cite{lacroix2018canonical}}      &      $\hat{\bm{h}},\hat{\bm{r}},\hat{\bm{t}}\!\in\!\mathbb R^{{d}}$, $\breve{\bm{h}},\breve{\bm{r}},\breve{\bm{t}}\!\in\!\mathbb R^{{d}}$       &                    $\left\langle \hat{\bm{h}}, \hat{\bm{r}}, \breve{\bm{t}} \right\rangle$ + $\left\langle \breve{\bm{h}}, \breve{\bm{r}}, \hat{\bm{t}} \right\rangle$              &  symmetric, anti-symmetric, asymmetric, inverse   \\ \hline
	\end{tabular} 
	\vspace{-10px}
\end{table*}

Given a set of observed (positive) triplets,
the goal of KGE is to learn low-dimensional vector representations of 
entities and relations so that 
the 
plausibility measured by $f(\bm{h}, \bm{r}, \bm{t})$ of observed triplets $(h,r,t)$s
are maximized while those of non-observed ones are minimized \cite{wang2017knowledge}.
To build a KGE model,
the most important thing is to design and choose a proper SF $f$,
which measures the triplets' plausibility based on embeddings.
Since different SFs have different strengths and weaknesses,
the choice of $f$ is critical for the KGE's performance \cite{wang2017knowledge,lin2018knowledge}.
A large amount of KGE models with popular SFs
follow the same framework (Alg.\ref{alg:general}) \cite{wang2017knowledge}
using stochastic gradient descent.
%Alg.\ref{alg:general} contains two main parts,
%i.e.  negative sampling (step~5) and embedding updates (step~6). 
%there are only positive triplets in the training data,
%negative sampling (step~5) is used to find 
%some negative triplets 
At step~5,
negative triplets are sampled from $\tilde{\mathcal{S}}_{(h,r,t)}$,
which contains all non-observed triplets for a current positive triplet $(h, r, t)$,
by some fixed distributions \cite{wang2014knowledge} or dynamic sampling schemes \cite{zhang2018nscaching}. 
%$\tilde{\mathcal{S}}_{(h,r,t)}$  is the set of all negative triplets for current positive triplet $(h, r, t)$,
%Then negative triplets are sampled from some fixed distributions \cite{wang2014knowledge} 
%or dynamic sampling schemes \cite{zhang2018nscaching}. 
Next,
the gradients
are computed 
based on the given SF and embeddings,
and
are used to update the model parameters (step~6).
Hinge loss \cite{bordes2013translating} and logistic loss \cite{yang2014embedding} are popularly used as $\ell$.
In this paper, we use the multi-class loss \cite{lacroix2018canonical}
since it
currently achieves the best performance and has little variance.
 
\begin{algorithm}[ht]
	\caption{Stochastic training of KGE \cite{wang2017knowledge}.}
	\label{alg:general}
	\small
	\begin{algorithmic}[1]
		\REQUIRE training set $\mathcal{S} = \{(h, r, t)\}$, scoring function $f$ and loss function $\ell$;
		
		\STATE initialize embeddings $\bm{e}, \bm{r}$ for each $e \in \mathcal E$ and $r \in \mathcal R$.
		\FOR{$i = 1, \cdots, T$}
		
		\STATE sample a mini-batch $\mathcal{S}_{\text{batch}} \subseteq \mathcal{S}$ of size $m$;
		% \STATE $\mathcal B^+\leftarrow \emptyset$, $\mathcal B^-\leftarrow \emptyset$
		\FOR{each $(h, r, t)\in\mathcal{S}_{\text{batch}}$}
		
		\STATE  sample $\tilde{m}$ negative triplets $\tilde{\mathcal{S}}_{(h,r,t)} \equiv 
		\{ (\tilde{h}_j, r, \tilde{t}_j) \}$
		for the positive triplet $(h, r, t)$; 
		
		\STATE update embedding parameters based on loss $\ell$ using selected positive and negative triplets;
		
		\ENDFOR
		\ENDFOR
		\RETURN embeddings of entities in $\mathcal{E}$ and relations in $\mathcal{R}$.
	\end{algorithmic}
\end{algorithm} 

Existing human-designed SFs mainly fall into three types:
\begin{itemize}[leftmargin=10px,itemsep = 5px]
	\item 
	\textit{Translational distance models (TDMs):}
	The translational approach exploits the distance-based SFs. 
	Inspired by the word analogy results in word embeddings \cite{bengio2013representation}, 
	the plausibility is measured based on the distance between two entities, 
	after a translation carried out by the relation. 
	In TransE \cite{bordes2013translating}, 
	the SF is defined by the (negative) distance between $\bm{h}+\bm{r}$ and $\bm{t}$, 
	i.e.  $f(\bm{h}, \bm{r}, \bm{t}) = -||\bm{h}+\bm{r}-\bm{t}||_1$. 
	Other TDMs-based SFs,
	e.g.,
	TransH \cite{wang2014knowledge}, TransR \cite{fan2014transition},
	enhance over TransE by introducing extra mapping matrices.

	\item 
	\textit{BiLinear models (BLMs):}
	SFs in this group exploit the plausibility of a triplet by the product-based similarity.
	%\footnote{+++ actually, all of them are bilinear?}
	%Bilinear models are the most well-known category in SMMs, 
	Generally,
	they share the form as 
	$f(\bm{h}, \bm{r}, \bm{t})=\bm{h}^\top\bm{R}\bm{t}$ 
	where $\bm{R}\!\in\!\mathbb R^{d\times d}$ is a matrix referring to the embedding of relation $r$ \cite{wang2017knowledge,wang2017multi}. 
	RESCAL \cite{nickel2011three} models the embedding of each relation by directly using $\bm{R}$.
	%which is big in size. 
	DistMult \cite{yang2014embedding} 
	overcomes the overfitting problem of RESCAL by constraining $\bm{R}$ to be diagonal.
	ComplEx \cite{trouillon2017knowledge} allows $\bm{R}$ and $\bm{h}, \bm{t}$ to be complex values,
	which enables handling asymmetric relations.
	HolE \cite{nickel2016holographic} uses a circular correlation to replace the dot product operation, 
	but is proven to be equivalent to ComplEx \cite{hayashi2017equivalence}. 
	Other variants like Analogy \cite{liu2017analogical}, SimplE \cite{kazemi2018simple}
	regularize the matrix $\bm{R}$ in different ways.
%	Analogy \cite{liu2017analogical} constrains $\bm{R}$ to be normal and commutative, 
%	and is implemented with a weighted combination of DistMult and ComplEx. 
%	Finally,
%	SimplE \cite{kazemi2018simple} uses two groups of embedding for each entity and relation to deal with the inverse relations. 
	
	\item 
	{
	\textit{Neural network models (NNMs):}
	Neural models aim to output the probability of the triplets based on neural networks 
	which take the entities' and relations' embeddings as inputs.
	MLP proposed in \cite{dong2014knowledge} and NTN proposed in \cite{socher2013reasoning} are representative neural models.
	Both of them use a large amount of parameters to combine entities' and relations' embeddings.
	ConvE \cite{dettmers2017convolutional} takes advantage of 
	convolutional neural network to increase the interaction among different dimensions of the embeddings.}
	
\end{itemize}

{
As proved in \cite{wang2017multi}, 
TDMs have less expressive ability than BLMs,
which further leads to their inferior empirical performance.
Based on the power of deep networks,
NNMs are also introduced for KGE.
However, 
due to the huge model complexity and increasing difficulty of training,
as well as the lack of domain-specific constraints,
their performance is still worse than BLMs
\cite{dettmers2017convolutional,lacroix2018canonical}.
Therefore,
we focus on BLMs in the sequel.
The most representative BLMs are listed in Tab.~\ref{tab:scofun}.}

\subsection{Automated machine learning (AutoML)}
\label{sec:automl}
Automated machine learning (AutoML) \cite{automl_book,quanming2018auto}
has recently exhibited its power
in easing the usage of and designing better machine learning models.
%the automated machine learning (AutoML) has become a hot topic recently. 
Basically,
AutoML can be regarded as a bi-level optimization problem
where we need to update model parameters by the training data sets
and tune hyper-parameters by the validation data sets.
Regarding the success of AutoML,
there are two important perspectives:
%\cite{feurer2015efficient,xie2017genetic,zoph2017neural,liu2018progressive}.
\begin{itemize}[leftmargin=8px]
	\item
	\textit{ Search space:}
	This helps to figure out important properties of the 
	underlying learning models and set up the search space for an AutoML problem.
	First,
	the space needs to be general enough
	to cover human wisdom as special cases.
	However,
	the space cannot be too general,
	otherwise searching in the space will be too expensive.
	\item
	\textit{ Search algorithm}:
	Unlike convex optimization,
	there is no universal and efficient optimization tools.
	%searching is generally inefficient and difficult.
	Once the search space is determined,
	efficient algorithms should be developed to search good points in the space.
\end{itemize}

We take NAS and HPO as examples.
The search space in NAS is spanned by network operations,
e.g., convolution with different sizes, skip-connections.
Various tailor-made algorithms,
such as 
reinforcement learning \cite{zoph2017neural}, 
evolution algorithms \cite{xie2017genetic},
and one-shot algorithms \cite{liu2018darts,yao2019differentiable},
have been proposed for efficient optimization.
For HPO,
Bayesian optimization \cite{feurer2015efficient,falkner2018bohb} is usually customized to search the space
made up by the hyper-parameters of the learning tools.

This paper is the first step towards automated embedding of knowledge graphs.
However,
such a step is not trivial
since previous AutoML methods used in NAS and HPO
cannot be directly applied to KGE.
The main problem is that we need to explore domain-specific properties 
in defining the search space and
designing efficient search algorithm
%here
to achieve effectiveness with less cost.

%\footnote{+++ pleas add $h^T$ and $t$ into Fig.1(e);
%	see example in the folder ``figure/example'' (this figure is extracted from your PPT)}

%Representative works are
%neural architecture search (NAS) \cite{zoph2017neural}
%%which helps to construct network architectures for convolutional networks and recurrent networks,
%and auto-sklearn \cite{feurer2015efficient}.
%%which aims at automatically finding hyper-parameters for configuring 
%%an ensemble of out-of-box classifiers implemented in sklearn \cite{scikit-learn}.
%These works have achieved exciting breakthroughs recently.

\begin{figure*}[ht]
	\centering
	\subfigure[DistMult.]
	{\includegraphics[height=3cm]{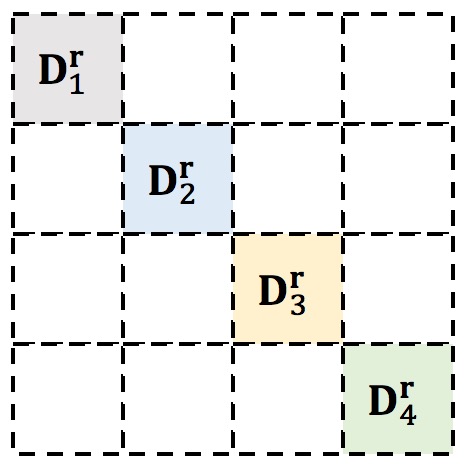}}\ 
	%	\quad 
	\subfigure[ComplEx.]
	{\includegraphics[height=3cm]{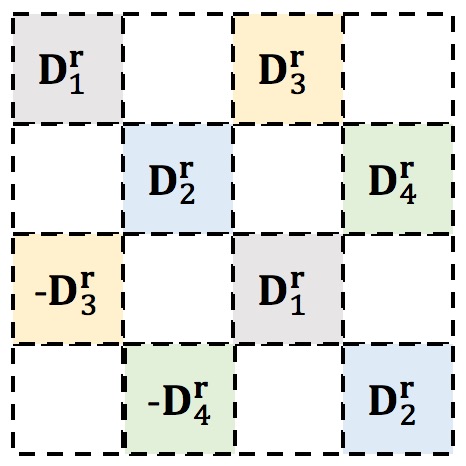}}\ 
	%	\quad
	\subfigure[Analogy.]
	{\includegraphics[height=3cm]{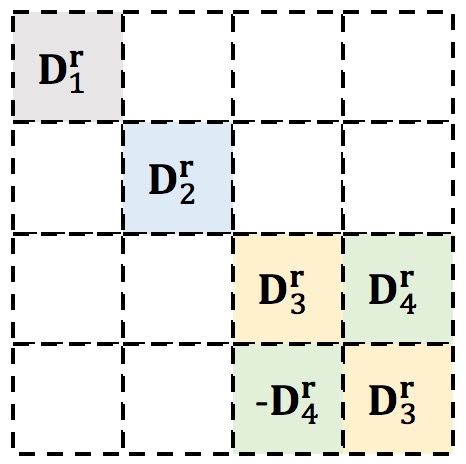}}\ 
	%	\quad
	\subfigure[SimplE.]
	{\includegraphics[height=3cm]{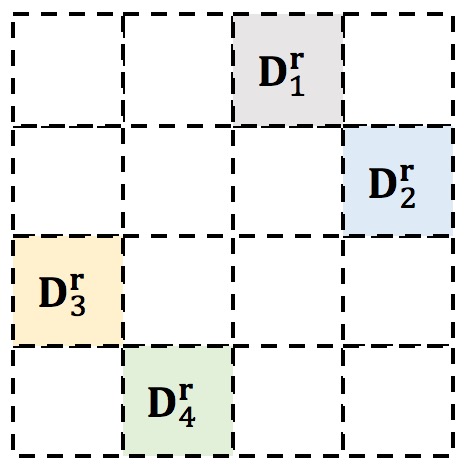}}
	\quad\quad
	\subfigure[AutoSF.]
	{\includegraphics[height=3cm]{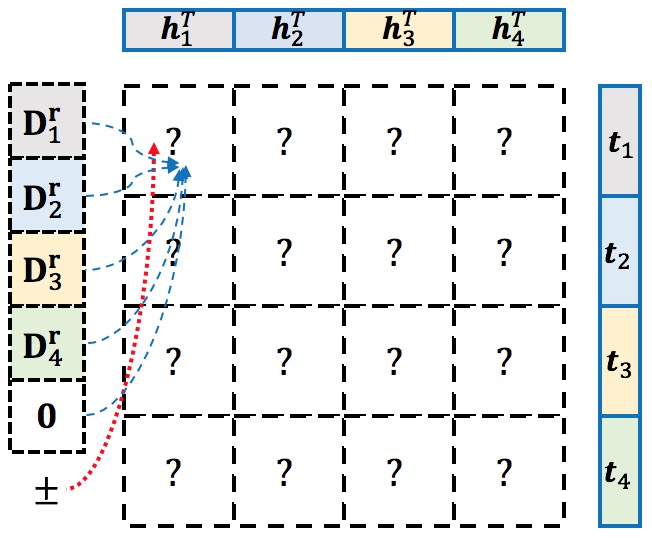}}
	\vspace{-5px}
	\caption{A graphical illustration of $\bm{R}$ for existing SFs in Tab.~\ref{tab:scofun}
		and the search space of AutoSF.
		Blank space is for zero matrix
		and $\bm{D}_i^{\bm{r}}=\diag{\bm{r}_i}, i = 1 \dots 4$.}
	\label{fig:graphsf}
	\vspace{-15px}
\end{figure*}

\section{The Search Problem} 
\label{sec:search}

As mentioned in Sec.\ref{sec:relworks}, 
new designs of SFs have continuously boosted
the performance of KGEs in recent years.
However, there is no absolute winner among the human-designed SFs.
Besides,
as different KGs usually exhibit distinct patterns in relations,
how to choose a proper SF to achieve good performance is non-trivial.
These raise one question:
\textit{can we automatically design a SF for a given KG with good performance guarantee}?
In this part,
we define AutoSF as a searching problem and make deep analysis on the search space based on KG properties
to address the question.

\subsection{AutoSF: Searching for SFs}
\label{ssec:searSFs}
%\footnote{$\surd$ too short}
Since SF is the crux to KGE
and different KG has distinct properties,
we are motivated to form the problem of designing new and better SFs
as a searching problem.
Specifically,
we define it as follows:

\begin{definition}[AutoML Problem]\label{def:autoSF}
Let $F(\bm{P}; g)$ be a KGE model (with indexed embeddings $\bm{P} = \{ \bm{h}, \bm{r}, \bm{t} \}$ and structure $g$),
$\mathcal M\left(F(\bm{P}; g), \mathcal{S}\right)$ 
measures the performance (the higher the better) of a KGE model $F$ on a set of triplets $\mathcal{S}$.
The problem of searching the SF is formulated as:
%%\vspace{-1px}
\begin{align}
g^*
&\in     
\arg\max\nolimits_{g \in \mathcal{G}}
\mathcal M\left(F(\bm{P}^*; g), \mathcal{S}_{\text{val}}\right)
\label{eq:autosf:l1}
\\
\;\;\text{s.t.}\;\; 
&\bm{P}^* 
=
\arg\max\nolimits_{\bm{P}} \mathcal M(F(\bm{P}; g), \mathcal{S}_{\text{tra}}),
\label{eq:autosf}
%%\vspace{-5px}
\end{align}
where $\mathcal{G}$ contains all possible choices of $g$,
$\mathcal{S}_{\text{tra}}$ and $\mathcal{S}_{\text{val}}$
denote training and validation data sets.
\end{definition}

Same as NAS \cite{xie2017genetic,zoph2017neural} 
and HPO \cite{feurer2015efficient,falkner2018bohb},
AutoSF is formulated as a bi-level optimization problem.
We firstly need to train the model to obtain $\bm{P}^*$ (converged model parameters) 
on the training set $\mathcal{S}_{\text{tra}}$ by \eqref{eq:autosf},
and then search for a better $g$
which is measured by the performance $\mathcal M$ on the validation set $\mathcal{S}_{\text{val}}$
by \eqref{eq:autosf:l1}.
However,
in this sequel
we can see the search space of $g$ and search strategy in AutoSF are
fundamentally different from previous AutoML works.
They are closely related to KGE's domain and new to the AutoML literature.

%While the AutoSF (Definition~\ref{def:autoKGE}) problem
%follows the same principles as the NAS \cite{xie2017genetic,zoph2017neural,baker2017designing} 
%and auto-sklearn \cite{feurer2015efficient},
%i.e.  both high-level configuration search and low-level model training are involved,
%they are fundamentally different.
%Specifically,
%in NAS the search space is made by the design choices of neural network architectures;
%in auto-sklearn the search space is comprised by hyper-parameters of out-of-box classifiers.
%The search space identified in AutoSF,
%i.e. choices of $g$ in Definition~\ref{def:autoKGE},
%is motivated by our unified representation
%over commonly used BLM-based SFs in Definition~\ref{def:unify}.
%Besides,
%we  clarify the importance of ensuring expressiveness in the search space,
%which are rooted in KGE's applications.
%These are all specific to KGE's domain and new to the AutoML literature.

\subsection{Search space: a unified representation of BLM SFs}
\label{ssec:unified}

{
To solve the AutoSF problem,
the first question is:
\textit{what is a good search space $\mathcal{G}$}?
%we need to figure out a reasonable search space $\mathcal{G}$ first.
As discussed in Sec.\ref{sec:automl},
the space can neither be too specific nor too general.
To motivate a good search space,
let us look at some commonly used SFs (Tab.~\ref{tab:scofun})
and dig out what are important properties of $g$.

As discussed in Sec.\ref{ssec:score},
the state-of-the-art performance of KGE models
is achieved by BLMs \cite{lacroix2018canonical,kazemi2018simple},
thus we limit our scope to them. 
RESCAL \cite{nickel2011three} is not considered 
since it does not have good scalability \cite{liu2017analogical,trouillon2017knowledge}
and neither empirically perform well.
The other models in Tab.~\ref{tab:scofun}
regularize the number of trainable parameters of
square matrix $\bm{R}\in\mathbb R^{d\times d}$ to be the same as entity embedding dimensions.
Therefore, we constrain the relation embedding size to be the same as the entity's
and learn different ways of mapping the relation embedding $\bm{r}\in\mathbb R^d$
into a square matrix $\bm{R}\in\mathbb R^{d\times d}$.
Besides, 
as the summary of relation types in Tab.~\ref{tab:scofun},
important properties are symmetric, anti-symmetric, asymmetric and inverse.
These are important properties of good SFs.
Thus, a search space should be able to handle the symmetric related properties.
In addition,
as will be discussed in Remark~\ref{rmk:reg},
different SFs in BLMs differ in their way of regularizing the square matrix $\bm R$.
Therefore, we are motivated to adaptively search how to regularize the relational matrix on different KGs.

To motivate such a space,
we can see that there are two main differences among these SFs.
\begin{itemize}[leftmargin=15px]
\item
The embedding can be either real or complex,
e.g. DistMult v.s. ComplEx.

\item
When embedding vectors are split,
different SFs combine them in distinct manners,
e.g., Analogy v.s. SimplE.
\end{itemize}

\begin{table*}[ht]
\caption{Common relations in KGs and resulting requirements on $f$, and search candidates in $g(\bm{r})$.}
\centering
\vspace{-8px}
\renewcommand{\arraystretch}{1.2}
\label{tab:comrel}
%	\small
\begin{tabular}{c | c | c | c}
	\hline
	                         common relations                          & requirements on $f$                                                            & requirements on $g(\bm{r})$       & examples from WN18/FB15K                \\ \hline
	                symmetric \cite{yang2014embedding}                 & $f(\bm{t},\bm{r},\bm{h}) = f(\bm{h}, \bm{r}, \bm{t})$                          & $g(\bm{r})^\top  =  g(\bm{r})$    & \textit{IsSimilarTo}, \textit{Spouse}   \\ \hline
	anti-symmetric \cite{trouillon2017knowledge,nickel2016holographic} & $f(\bm{t},\bm{r},\bm{h}) = -f(\bm{h}, \bm{r}, \bm{t})$                         & $g(\bm{r})^\top  = -g(\bm{r})$    & \textit{LargerThan}, \textit{Hypernym}  \\ \hline
	           general asymmetric \cite{liu2017analogical}             & $f(\bm{t},\bm{r},\bm{h})\neq f(\bm{h}, \bm{r}, \bm{t})$                        & $g(\bm{r})^\top  \neq  g(\bm{r})$ & \textit{LocatedIn}, \textit{Profession} \\ \hline
	                 inverse \cite{kazemi2018simple}                   & $f(\bm{t},\bm{r},\bm{h}) = f(\bm{h},\bm{r}',\bm{t}), \mathbf r\neq \mathbf r'$ & $g(\bm{r})^\top  =  g(\bm{r}')$   & \textit{Hypernym}, \textit{Hyponym}     \\ \hline
\end{tabular}
\vspace{-15px}
\end{table*}

\subsubsection{Dealing with complex embeddings}
A complex vector $\bm v\!\in\!\mathbb C^d$ with $\bm v = \bm v_{re}\!+\!i\bm v_{im}$
is composed of a real part $\!\bm v_{re}\!\in\!\mathbb R^d\!$ and an imaginary part $\bm v_{im}\!\in\!\mathbb R^d$.
To deal with the complex embeddings, we can use $2d$-dimensional real vector $[\bm v_{re}, \!\bm v_{im}]$ to represent the $d$-dimensional complex vector $\bm v$ \cite{trouillon2017knowledge,balavzevic2019tucker}.
Let the complex embedding $\!\bm{h}\!=\!\bm{h}_{re}\!+\!i\bm{h}_{im}$,
where $\bm{h}_{re}, \bm{h}_{im}\in\mathbb R^d$ (same for $\bm{r},\bm{t}$),
then ComplEx can be expressed  as 
\begin{align}
\text{Re}\left(\left\langle\bm{h}, \bm{r}, \text{conj}(\bm{t}) \right\rangle \right) 
&=\left\langle \bm{h}_{re},\bm{r}_{re}, \bm{t}_{re}\right\rangle 
+\left\langle \bm{h}_{im},\bm{r}_{re}, \bm{t}_{im}\right\rangle  \nonumber\\
+ & \left\langle \bm{h}_{re},\bm{r}_{im}, \bm{t}_{im}\right\rangle
-\left\langle \bm{h}_{im},\bm{r}_{im}, \bm{t}_{re}\right\rangle.
\label{eq:complex}
\end{align}
Similarly, DistMult \cite{yang2014embedding} with $2d$-dimensional embeddings can also be denoted by $[\bm v_{re}, \bm v_{im}]$ and represented as two parts
\begin{align}
\left\langle\bm{h}, \bm{r}, \bm{t} \right\rangle
&=\left\langle \bm{h}_{re},\bm{r}_{re}, \bm{t}_{re}\right\rangle 
+\left\langle \bm{h}_{im},\bm{r}_{im}, \bm{t}_{im}\right\rangle.
%\vspace{-3px}
\label{eq:distmult}
\end{align}

\subsubsection{Dealing with different splits}
To make the training parameters consistent,
we also use $2d$-dimensional real valued embeddings to represent Analogy and SimplE.
As given in Tab.~\ref{tab:scofun},
embeddings in Analogy \cite{liu2017analogical} are split into a real part $\hat{\bm h}\in\mathbb R^d$,
and a complex part $\breve{\bm h}$,
which can be denoted as a concatenated real vector $[\breve{\bm h}_{re}, \breve{\bm h}_{im}] \in \mathbb R^{d}$
in the similar way as ComplEx. And the SF is split as
\begin{equation}
\left\langle \hat{\bm{h}}, \hat{\bm{r}}, \hat{\bm{t}}\right\rangle + \text{Re}\left( \left\langle  \breve{\bm{h}}, \breve{\bm{r}}, \text{conj}(\breve{\bm{t}})\right\rangle \right).
\label{eq:analogy}
\end{equation}
In SimplE \cite{kazemi2018simple}, 
two independent embedding vectors $\hat{\bm h}\in\mathbb R^d$ and $\breve{\bm h}\in\mathbb R^d$
are used to represent each entity and relation.
The resulting SF becomes
\begin{equation}
\left\langle \hat{\bm{h}}, \hat{\bm{r}}, \breve{\bm{t}} \right\rangle + \left\langle \breve{\bm{h}}, \breve{\bm{r}}, \hat{\bm{t}} \right\rangle.
\label{eq:simple}
\end{equation}

\subsubsection{The unified representation}
In order to deal with the two different partitions,
i.e. ComplEx v.s. DistMult 
and Analogy v.s. SimplE,
we split embedding 
$\bm{h} \in \mathbb{R}^d$ as $\bm{h}  = [ \bm{h}_1; \bm{h}_2; \bm{h}_3; \bm{h}_4]$
(same for $\bm{r}$ and $\bm{t}$)
to cover \eqref{eq:complex}, \eqref{eq:distmult}, \eqref{eq:analogy} and \eqref{eq:simple}.
Note that any splits $k$ (with $k\geq 4$ and $k$ is even) can be used to cover the SFs in Tab.~\ref{tab:scofun}.
We take $k=4$ in order to ensure a tractable search space.
%The transformation of each SF is then summarized in the last column of 
%\footnote{$\surd$ explain the caption for the table}
%Tab.~\ref{tab:scofun}.
The transformation of each SF is then summarized as:
%\begin{itemize}[leftmargin=10px]
%\item DistMult:
%$\left\langle  \bm{h}_1, \bm{r}_1, \bm{t}_1\right\rangle 
%+ \left\langle\bm{h}_2, \bm{r}_2, \bm{t}_2\right\rangle 
%+ \left\langle\bm{h}_3, \bm{r}_3, \bm{t}_3\right\rangle
%+ \left\langle\bm{h}_4, \bm{r}_4, \bm{t}_4\right\rangle$;
%
%\item ComplEx:
%$\left\langle \bm{h}_1, \bm{r}_1, \bm{t}_1\right\rangle 
%+ \left\langle \bm{h}_1, \bm{r}_3, \bm{t}_3\right\rangle 
%+ \left\langle\bm{h}_3, \bm{r}_1, \bm{t}_3\right\rangle 
%- \left\langle\bm{h}_3, \bm{r}_3, \bm{t}_1\right\rangle
%+ \left\langle\bm{h}_2, \bm{r}_2, \bm{t}_2\right\rangle
%+ \left\langle\bm{h}_2, \bm{r}_4, \bm{t}_4\right\rangle 
%+ \left\langle\bm{h}_4, \bm{r}_2, \bm{t}_4\right\rangle
%- \left\langle\bm{h}_4, \bm{r}_4, \bm{t}_2\right\rangle$;
%
%\item Analogy:
%$\left\langle \bm{h}_1, \bm{r}_1, \bm{t}_1\right\rangle 
%+ \left\langle\bm{h}_2, \bm{r}_2, \bm{t}_2\right\rangle
%+ \left\langle\bm{h}_3, \bm{r}_3, \bm{t}_3\right\rangle 
%+ \left\langle\bm{h}_3, \bm{r}_4, \bm{t}_4\right\rangle 
%+ \left\langle\bm{h}_4, \bm{r}_3, \bm{t}_4\right\rangle
%- \left\langle\bm{h}_4, \bm{r}_4, \bm{t}_3\right\rangle$;
%
%\item SimplE / CP:
%$\left\langle \bm{h}_1, \bm{r}_1, \bm{t}_3\right\rangle 
%+ \left\langle\bm{h}_2, \bm{r}_2, \bm{t}_4\right\rangle
%+ \left\langle\bm{h}_3,\bm{r}_3,\bm{t}_1\right\rangle
%+ \left\langle\bm{h}_4, \bm{r}_4, \bm{t}_2\right\rangle$.
%\end{itemize}
{
\small
\begin{align*}
\text{DistMult}\!\!:
\!f(\mathbf{h}, \!\mathbf{r}, \!\mathbf{t}) 
& \!\!=\!\! \left\langle  \mathbf{h}_1, \!\mathbf{r}_1, \!\mathbf{t}_1\right\rangle 
\!\!+\!\! \left\langle\mathbf{h}_2, \!\mathbf{r}_2,\! \mathbf{t}_2\right\rangle
\!\!+\!\! \left\langle\mathbf{h}_3, \!\mathbf{r}_3, \!\mathbf{t}_3\right\rangle
\!\!+\!\! \left\langle\mathbf{h}_4, \!\mathbf{r}_4, \!\mathbf{t}_4\right\rangle\!,
\\
\text{ComplEx}\!\!:
\!f(\mathbf{h}, \!\mathbf{r},\! \mathbf{t})
& \!\!=\!\!  \left\langle \mathbf{h}_1, \!\mathbf{r}_1,\! \mathbf{t}_1\right\rangle 
\!\!+\!\! \left\langle \mathbf{h}_1, \!\mathbf{r}_3, \!\mathbf{t}_3\right\rangle 
\!\!+\!\! \left\langle\mathbf{h}_3, \!\mathbf{r}_1,\! \mathbf{t}_3\right\rangle 
\!\!-\!\! \left\langle\mathbf{h}_3, \!\mathbf{r}_3, \!\mathbf{t}_1\right\rangle
\\
&\!\!+\! \left\langle\mathbf{h}_2, \!\mathbf{r}_2, \!\mathbf{t}_2\right\rangle
\!\!+\!\! \left\langle\mathbf{h}_2, \!\mathbf{r}_4, \!\mathbf{t}_4\right\rangle 
\!\!+\!\! \left\langle\mathbf{h}_4, \!\mathbf{r}_2, \!\mathbf{t}_4\right\rangle
\!\!-\!\! \left\langle\mathbf{h}_4, \!\mathbf{r}_4, \!\mathbf{t}_2\right\rangle\!,
\\
\text{Analogy}\!\!:
\!f(\mathbf{h},\! \mathbf{r}, \!\mathbf{t})
& \!\!=\!\! \left\langle \mathbf{h}_1, \!\mathbf{r}_1, \!\mathbf{t}_1\right\rangle 
\!\!+\!\!\left\langle\mathbf{h}_2, \!\mathbf{r}_2, \!\mathbf{t}_2\right\rangle
\!\!+\!\! \left\langle\mathbf{h}_3, \!\mathbf{r}_3, \!\mathbf{t}_3\right\rangle 
\!\!+\!\! \left\langle\mathbf{h}_3,\! \mathbf{r}_4, \!\mathbf{t}_4\right\rangle 
\\
&
\!\!+\! \left\langle\mathbf{h}_4, \!\mathbf{r}_3, \!\mathbf{t}_4\right\rangle
\!\!-\!\! \left\langle\mathbf{h}_4, \!\mathbf{r}_4, \!\mathbf{t}_3\right\rangle\!,
\\
\text{SimplE}\!:
\!f(\mathbf{h},\! \mathbf{r}, \!\mathbf{t})
& \!\!=\!\! \left\langle \mathbf{h}_1, \!\mathbf{r}_1, \!\mathbf{t}_3\right\rangle 
\!\!+\!\! \left\langle\mathbf{h}_2, \!\mathbf{r}_2, \!\mathbf{t}_4\right\rangle
\!\!+\!\! \left\langle\mathbf{h}_3,\!\mathbf{r}_3,\!\mathbf{t}_1\right\rangle
\!\!+\!\! \left\langle\mathbf{h}_4, \!\mathbf{r}_4, \!\mathbf{t}_2\right\rangle\!.
\end{align*}
}

\vspace{-10px}

Based on above formulations,
all the scoring functions can be formed as $f(\bm h, \bm r, \bm t) = \bm h^\top \bm R \bm t$.
%these SFs are equivalent to 
%the bilinear function $\bm{h}^\top \bm{R} \bm{t}$ 
%with a different form of a square relation matrix $\bm{R}$.
%\footnote{+++ $\bm{R}_i^{\bm{R}}$ is better?}
Let $\mathbf{D}^{\bm{r}}_i \!=\! \diag{\bm{r}_i}$ for $i \in \{1$,$2$,$3$,$4 \}$,
the forms of  $\bm{R}$ for these SFs can be graphically represented as Fig.~\ref{fig:graphsf}.
In this way, we can see that the main difference between the four SFs is 
their way of filling the $4\times 4$ block
matrix (see Fig.~\ref{fig:graphsf}(e)).
Based on such a pattern,
we identify
the search space of BLM-based SFs in Definition~\ref{def:unify}.

}

\begin{definition}[Search space $\mathcal{G}$] \label{def:unify}
%\footnote{+++ what is the dimension of $\bm{R}$?}
%\footnote{+++ $\bm{h}^{\top} \bm{R} \bm{t}$ has been mentioned previously,
%	the key is the structure in $\bm{R}$,
%	be more careful on the statement.}
Let $g\left( \bm{r} \right)$ return
a $4 \times 4$ block matrix,
of which the elements in each block is given by 
$\left[  g\left( \bm{r} \right) \right]_{ij} = \diag{\bm{a}_{ij}}$
where 
$\bm{a}_{ij} \in \{ \mathbf{0}, \pm \bm{r}_1, \pm \bm{r}_2, \pm \bm{r}_3, \pm \bm{r}_4 \}$
for $i,j \in \{1, 2, 3, 4\}$.
Then,
SFs can be represented by
\begin{align*}
\vspace{-1px}
f_{\text{unified}}(\bm{h},\bm{r},\bm{t})  
= \sum\nolimits_{i,j}\left\langle\bm{h}_i, \mathbf a_{ij}, \bm{t}_j\right\rangle = \bm{h}^{\top} g\left( \bm{r} \right) \bm{t}.
\label{eq:uni}
\vspace{-1px}
\end{align*}
\end{definition}

\begin{remark}[Searching to regularize BLMs]
\label{rmk:reg}
Note that the SFs shown in Fig.~\ref{fig:searchedsf} constrain the relation matrix $\bm R$ in different forms,
which can be regarded as different regularization schemes.
Viewed in this way, AutoSF aims to search how to regularize the relational matrix 
that can adapt to different relation properties in different  KGs.
In addition,
the data dependent regularization cannot be easily formed as a constraint in training procedure,
which motivates us to use AutoML to search based on validation sets performance.
%\footnote{$\surd$
%discuss the 4th reviewer's
%point here.}
%\footnote{$\surd$ check Sec.\ref{ssec:casestufy}.
%	experiments there can explain point here}
%\footnote{$\surd$ explain such data dependent regularization
%	can only be searched from validation set (i.e.  by AutoML).}
\end{remark}

{
\begin{remark}[General Search Space]
Due to the recent success of deep networks \cite{goodfellow2016deep}
and the approximation ability of multilayer perceptron (MLP) \cite{csaji2001approximation},
one may want to use a MLP as $\mathcal{F}$ for \eqref{eq:autosf}.
However,
the design of the MLP is also a searching problem,
which is 
%and
%searching the network structure is 
very time-consuming \cite{zoph2017neural}.
Besides, an arbitrarily large MLP will lead to an extremely large space.
%(see details in Appendix~\ref{app:mlp}). 
As verified in \cite{yao2019searching},
the general approximator MLP is a bad choice for NAS 
and performs worse than those using reinforcement learning \cite{zoph2017neural}.
\end{remark}
}

\section{The Search Strategy}
\label{sec:progred}
%In Sec.\ref{sec:search}, we have defined the AutoSF problem,
%of which the crux is searching $g$ for SFs.
Here,
we propose an efficient search strategy to address the AutoSF problem
based on domain-specific properties in KGs.

\subsection{Challenges in algorithm design}
\label{ssec:challange}

Same as the other AutoML problems,
the search problem of AutoSF is black-box,
the search space is huge,
and each step in the search is very expensive since both model training and evaluation should be involved.
These problems have previously been touched by algorithms such as reinforcement learning \cite{zoph2017neural}, 
Bayes optimization \cite{feurer2015efficient}
and genetic programming \cite{xie2017genetic}.
However,
they are not a good choice here
since we have domain-specific problems,
i.e.  expressiveness and invariance,
 in KGE,
which are more challenging.

\subsubsection{Expressiveness}
\label{ssec:expre}

It is clear that
not all SFs $g \in \mathcal{G}$ (from Definition~\ref{def:unify}) are equally good.
Expressiveness (Definition~\ref{def:exp}), 
which means $f$ should be better able to handle common relations in KGs ,
is of big concern for SFs.
Their consequent requirements on $f$ and $g(\bm{r})$ are summarized in Tab.~\ref{tab:comrel}.

\begin{definition}[Expressiveness \cite{trouillon2017knowledge,kazemi2018simple,wang2017multi}] \label{def:exp}
	If $f$ can handle 
	symmetric,
	anti-symmetric,
	general asymmetric,
	and inverse relations,
	then $f$ is expressive.
\end{definition}

\begin{figure*}[ht]
	\centering
	\subfigure[SimplE.]
	{\includegraphics[width=0.32\columnwidth]{figure/simple}}
	\subfigure[sym.]
	{\includegraphics[width=0.32\columnwidth]{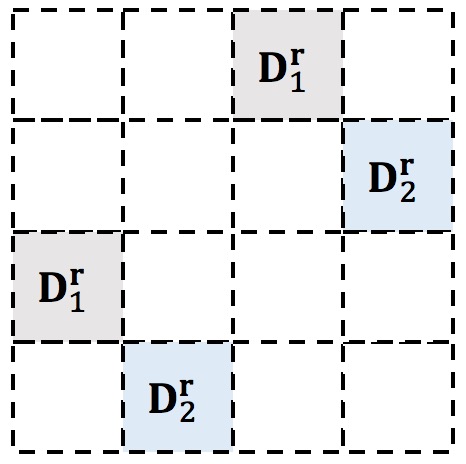}}
	\subfigure[skew-sym.]
	{\includegraphics[width=0.32\columnwidth]{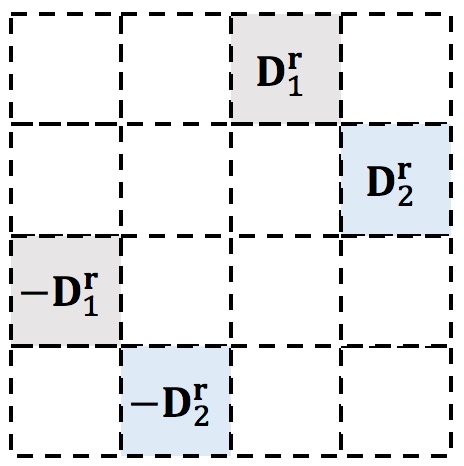}}
	\subfigure[permute $\bm{h}_i$ and $\bm{t}_i$.]
	{\includegraphics[width=0.32\columnwidth]{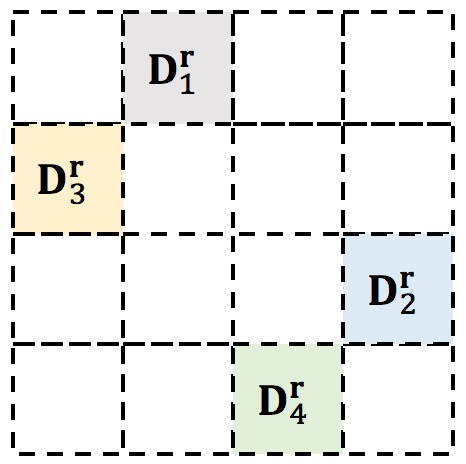}}
	\subfigure[permute $\bm{r}_i$.]
	{\includegraphics[width=0.32\columnwidth]{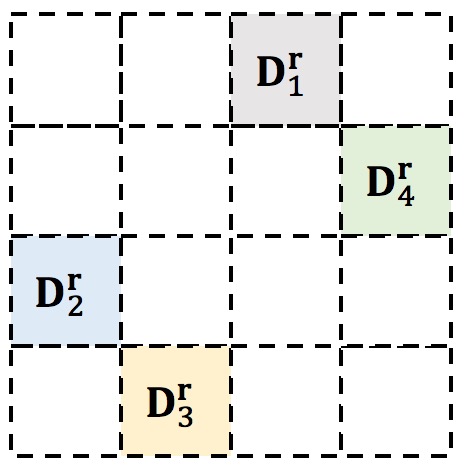}}
	\subfigure[flip sign.]
	{\includegraphics[width=0.32\columnwidth]{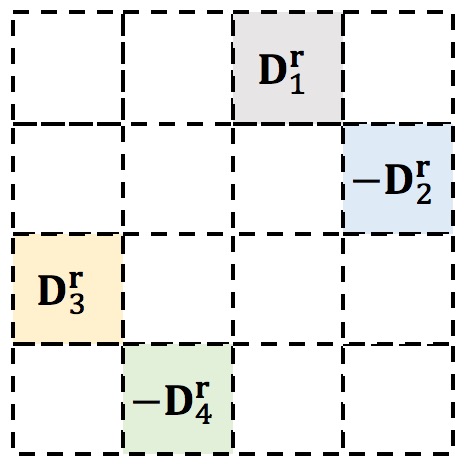}}
	
	\vspace{-6px}
	\caption{Illustration of \textit{Invariance} and \textit{Expressiveness}.
		(a) SimplE model;
		(b) assign $\bm{r}_3=\bm{r}_1, \bm{r}_4 = \bm{r}_2$;
		(c) assign $\bm{r}_3=-\bm{r}_1, \bm{r}_4 = -\bm{r}_2$;
		(d) permute $[\bm{h}_1; \bm{h}_2; \bm{h}_3; \bm{h}_4]$ into $[\bm{h}_1; \bm{h}_3; \bm{h}_2; \bm{h}_4]$ and do the same for $\bm t$;
		(e) permute $[\bm{r}_1; \bm{r}_2; \bm{r}_3; \bm{r}_4]$ into $ [\bm{r}_1; \bm{r}_4; \bm{r}_2; \bm{r}_3]$;
		(f) flip the signs of $\bm{r}_2$ and $\bm{r}_4$.
	}
	\label{fig:invariance}
	\vspace{-10px}
\end{figure*}

To ensure that
$f$ can handle those common relations,
we propose Proposition~\ref{pr:expre}.

\begin{prop}\label{pr:expre}
	If $g(\bm{r})$ can 
	be symmetric for some $\bm{r} \! \in \! \mathbb{R}^d$,
	i.e. $g(\bm{r})^\top \! = \! g(\bm{r})$,	
	and skew-symmetric
	for some $\bm{r}' \! \in \! \mathbb{R}^d$,
	i.e. $g(\bm{r}')^\top \! = \! -g(\bm{r}')$.
	Then
	the formulated $f$  is expressive. (Proofs in Appendix \ref{app:proof:expre}).
	%\vspace{-3px}
\end{prop}

With this Proposition
and to avoid trivial solutions,
we introduce the following constraints on $g$:

\begin{itemize}[leftmargin=25pt,topsep=0pt,parsep=0pt,partopsep=0pt]
	\item[(C1).] 
	$g(\bm{r})$ can be  symmetric 
	%	($g(\bm{R})^\top=g(\bm{R})$) 
	with proper $\bm{r}$
	and skew-symmetric
	%	($g(\mathbf{r'})^\top=-g(\mathbf{r'})$)
	with proper $\bm{r}'$.
	\item[(C2).] $g(\bm{r})$ has no zero rows/columns, 
	covers all $\bm{r}_1$ to $\bm{r}_4$, 
	and has no repeated rows/columns.
\end{itemize}
For (C1), 
the symmetric property of $g(\bm{r})$ determines what kind of relation the given SF can model
based on Proposition~\ref{pr:expre}.
For (C2), if there are zero rows/columns in $g$, the corresponding embedding dimensions will be useless.
It means that these dimensions will never be optimized during training.
%In short, (C2) requires the relation matrix $g(\bm{r})$ to be full-rank to utilize each embedding dimension well.

The above constraints are important 
for finding potentially good candidate $g \in \mathcal{G}$,
and they play a key role in filtering out the bad $g$'s for  the design of an efficient search algorithm. 
%\footnote{+++ I think it is better put Sec.\ref{ssec:expre} here.
%	Expressiveness and Invariance are of equal important to the search space.}
As in Definition~\ref{def:exp} and Proposition~\ref{pr:expre},
we need to deal with Constraint~(C1) for expressiveness.
It is challenging since 
$g$ only represents a structure,
however the exact check of (C1) relies on the values in $\bm{r}$,
which are unknown in advance.
Fortunately, we can check it by value assignment.
Take the SF in Fig.~\ref{fig:invariance}(a) for example.
We can see that $g(\bm{r})$ can be symmetric by assigning $\bm{r}_3=\bm{r}_1$
and $\bm{r}_4=\bm{r}_2$ as in 
Fig.~\ref{fig:invariance}(b),
and skew-symmetric by setting $\bm{r}_3 = -\bm{r}_2$
and $\bm{r}_4 = -\bm{r}_1$ like in Fig.~\ref{fig:invariance}(c).
This is the key idea of addressing expressiveness.

\subsubsection{Invariance}
As defined in Sec.\ref{ssec:unified},
the embeddings are split into 4 parts, i.e.
$\bm{r} = [\bm{r}_1; \bm{r}_2; \bm{r}_3; \bm{r}_4]$.
Prior to the embedding training,
permuting the $\bm{r}_i$'s will lead to equivalent structure 
since``1,2,3,4" here are only identity of each component
and these components are equivalent at this stage.
For example, we can permute $\bm{r} = [\bm{r}_1; \bm{r}_2; \bm{r}_3; \bm{r}_4]$ into $\bm{r} = [\bm{r}_2; \bm{r}_1; \bm{r}_3; \bm{r}_4]$.
Even though $\bm{r}_1$ and $\bm{r}_2$ change their position,
the learned embedding could be the same by changing corresponding values after training.
Therefore, the structure of SFs is invariance to permutation of $\bm{r}_i$'s.
Similarly,
since $\bm{h}$ and $\bm{t}$ share the same embedding parameters $\bm{e}$, 
the generated SFs are also equivalent by simultaneously permuting the $\bm{h}_i$'s and $\bm{t}_i$'s.
Moreover, if we flip the signs of some $\bm{r}_i$,
we can learn equal embeddings by flipping the true value of  those $\bm{r}_i$ after training.
In summary,
there exist three kinds of invariance: 
permuting entity embedding $\bm{h}_i$'s and $\bm t_i$'s,
permuting relation embedding $\bm{r}_i$'s, 
and flipping signs.
%\footnote{$\surd$ what are $[e_1, \cdots, e_4]$ in Fig.~\ref{fig:invariance}(d-f)?
%	You'd better change them with $\bm h$ and $\bm t$.
%	$\bm{e}$ is not commonly used in this paper}
%\footnote{$\surd$ also you can remind readers,
%	invariance means after model training $\bm{h}$ and $\bar{\bm{h}}$ (same for $\bm{r}$ and $\bm{t}$)  
%	$\bm{h}^{\top} g_1(\bm{r}) \bm{t} = \bar{\bm{h}}^{\top} g_2(\bar{\bm{r}}) \bar{\bm{t}}$.}
The example of three cases are given in Fig.~\ref{fig:invariance}(d-f).
Let $\bm h, \bm r,\bm t$ be the embeddings of the SF $g_1$
 and $\bar{\bm h}, \bar{\bm r},\bar{\bm t}$ be the embedding of another SF $g_2$,
 and $g_2$ is formed through the invariance changes of $g_1$.
 Then we will have  $\bm{h}^{\top} g_1(\bm{r}) \bm{t} = \bar{\bm{h}}^{\top} g_2(\bar{\bm{r}}) \bar{\bm{t}}$
 after the model training.
 Therefore,
it is tedious to train and evaluate the equivalents once we know the performance of one SF among them.

\subsection{Progressive greedy search}
\label{ssec:greedy}
%%\vspace{-5px}

As in Sec.\ref{ssec:unified},
adding one more block into $g$ indicates adding
one more nonzero multiplicative term into $f$,
namely
\begin{equation}
f^{b + 1} = f^b + s\left\langle \bm{h}_{i}, \bm{r}_{j}, \bm{t}_{k} \right\rangle,
\label{eq:progressive}
\end{equation}
where $s\!\in\!\{\pm 1 \}$ and $i,j,k\!\in\!\{1,2,3,4\} $.
In order to search efficiently, 
we propose a progressive greedy algorithm based on the inductive rule \eqref{eq:progressive},
which can significantly cut down the search space in a stage-wise manner.
The intuition of using \eqref{eq:progressive} to progressively generate SFs it to
gradually adjust the relation matrix $g(\bm{r})$.
However,
greedy search usually leads to sub-optimal solutions \cite{tropp2004greed},
which can be more serious when faced with the expressive and invariance challenges in AutoSF. 
Therefore,
we enhance the greedy search with a filter and a predictor
to specifically deal with  the expressiveness and invariance
discussed in Sec.\ref{ssec:challange}.

\begin{algorithm}[ht]
	\caption{Progressive greedy search algorithm.}
	\label{alg:greedy}
	\small
	\begin{algorithmic}[1]
		\REQUIRE $B$: number of nonzero blocks in $g$, a learnable predictor $\mathcal P$;
		\FOR{$b$ in $4,6,8,\cdots, B$}
		\REPEAT	 \label{step:gen-start}
		\STATE randomly select a top-$K_1$ model $f^{b-2} \in \mathcal T^{b-2}$; 
		\STATE generate 6 integers $i_1, j_1, k_1, i_2, j_2, k_2 \in \{1, 2, 3, 4\}$ and $s_1, s_2$ from $\{ \pm 1 \}$, 
		and form
		$		f^b \leftarrow f^{b-2} 
		+ s_1 \left\langle \bm{h}_{i_1}, \bm{r}_{j_1}, \bm{t}_{k_1}  \right\rangle 
		+ s_2 \left\langle \bm{h}_{i_2}, \bm{r}_{j_2}, \bm{t}_{k_2} \right\rangle$;
		\label{step:gen}
		
		\STATE \textbf{if} $f^b$ satisfies filter $\mathcal Q$ \textbf{then} $\mathcal{H}^b \leftarrow \mathcal{H}^b \cup \left\lbrace f^b\right\rbrace $;
		%		\STATE generate a candidate $f^{b}$ based on top $K_1$ SFs in $\mathcal T^{b-2}$.
		%		\label{step:gen}
		%		\STATE use the \textit{filter} $\mathcal{Q}$ to judge whether $f^{b}$ can be added in $\mathcal H^{b}$ 
		\label{step:filter}
		\UNTIL{$\left|\mathcal H^{b}\right|=N$}  \label{step:gen-end}
		%			\STATE generate a set $\mathcal H^{b_i}$ of $N$ candidates $f^{b_i}$ based on $\mathcal T^{b_i-2}$ (details in Appendix);
		%			\STATE filter out equivalent candidates in $\mathcal H^{b_i}$ and $\mathcal T^{b_i}$;
		%			\label{step:filter}
		
		{
		\STATE select top-$K_2$ $f^{b}$s in $\mathcal H^{b}$ based on the \textit{predictor} $\mathcal P$;
		\label{step:predict}
		
		\STATE \textit{train} embeddings with the selected $f^b$s;
		
		\STATE \textit{evaluate} the obtained embedding from the selected $f^b$s;	
		}
		
		\STATE $\mathcal{T}^{b} \leftarrow$ add and record $f^{b}$ and their performance;
		\label{step:record}
		%			\STATE $\mathcal{C}^{b_i} \leftarrow$ keep $f^{b_i}$ with top-$K_1$ performance in $\mathcal T^{b_i}$;
		\label{step:top}
		\STATE update the predictor $\mathcal P$ with records in $\mathcal T$. 
		\label{step:update} 
		\ENDFOR
		\RETURN desired SFs in $\mathcal{T}^B$.
	\end{algorithmic}
\end{algorithm}

\subsubsection{Complete procedures}
Alg.\ref{alg:greedy} shows our progressive greedy algorithm.
As in Definition~\ref{def:unify},
let the number of 
%\footnote{$\surd$+++ remind readers what is non-zero blocks?}
nonzero blocks in $g$ be $B$
and the SF in this group be $f^B$.
% and SF with $B$ nonzero multiplicative terms be $f^B$.
%The basic idea is simple.
%\footnote{
%	+++ basically we hope that adding more blocks can help fine tune the structure, 
%	this can work as another motivation of using greedy.
%}
%Since there are only a few candidates for $f^4$ (details in Appendix~\ref{app:f4}) 
%under Constraints~(C2),
The idea of progressive search is that
given the desired $B$,
we start from small blocks $b$ and then gradually add in more blocks until $b = B$.
Thus, 
we can greedily generate candidates based on the top SFs in $\mathcal T^{b-2}$ at step~\ref{step:gen-start}-\ref{step:gen-end}
%for each $b$
to reduce search space.
Specifically, we greedily pick up the top-$K_1$ $f^{b-2}$ in the previously evaluated models in $\mathcal T^{b-2}$.
$N$ candidates
will then be generated by adding two more multiplicative term in step~\ref{step:gen}
to deal with Constraint (C1)
since adding one block each step will result in simply lying on the diagonal.
All the candidates are generated from $b=4$ 
and are checked through the Filter $\mathcal Q$ (see Sec.\ref{ssec:filter}) to guarantee Constraint~(C2)
and avoid training equivalents.
%Note that,
%to deal with Constraint (C1),
%we increase $b$ by $2$ in each step to avoid trivially lying on the diagonal,
%and start from $b=4$ to keep consistent with (C2).
%A set $\mathcal H^{b}$ with $N$ candidate $f^{b}$'s is collected through the 
%\footnote{$\surd$ explicitly write Alg.3 in Alg.2.}
%Alg.\ref{alg:candgen}.
%The generation process (details in Appendix) is greedily based on top SFs in $\mathcal T^{b_i-2}$ and enhanced with a filter to avoid training on redundant SFs.
%filter out equivalent candidates 
% in $\mathcal H^{b_i}$
%and those that are equivalent to the visited SFs in $\mathcal T^{b_i-2}$.
%Candidates not satisfying (C2) should also be filtered out.
%The filter $\mathcal M$ can help us avoid training a lot of poor and redundant SFs in $\mathcal H^{b_i}$.
Next,
we use the predictor $\mathcal P$ (see Sec.\ref{ssec:pred}) to further select $K_2$ promising candidates,
which will then be trained and evaluated using Alg.\ref{alg:general},
%based on their structures $g$
in 
%\footnote{+++ step is wrong.}
step~\ref{step:predict} of Alg.\ref{alg:greedy}.
The training data for $\mathcal P$ is gradually collected with the trained SFs in $\mathcal{T}=\mathcal T^4\cup\mathcal T^6\cup \cdots$ at step~\ref{step:record}.

\subsubsection{Invariance - Using a filter}
\label{ssec:filter}

The filter $\mathcal Q$ we used in Alg.\ref{alg:greedy} has two functions:
1) deal with Constraint~(C2) 
and 2) remove equivalent structures due to invariance.
Constraint~(C2) is easy to check,
given the structure of $g$,
we can directly map it into a $4\times 4$ substitute matrix 
and use $\{0,\pm1, \pm2, \pm2, \pm4\}$
to represent 
$[g(\bm{r})]_{ij}\in \{\mathbf 0, \pm\bm{r}_1, \pm \bm{r}_2, \pm \bm{r}_3, \pm\bm{r}_4\}$.
Then checking requirements in (C2) is a trivial task,
i.e.  checking if the $4\times 4$ substitute matrix 
satisfies the Constraint~(C2).
%has zero or repeated rows/columns,
%and covers all the $\bm r_i$'s.

For the invariance,
once a candidate $f^b$ fulfilling Constraint~(C2) is generated, we use the invariance property to generate a set of equivalents $\mathcal G_{f^b}$.
Specifically,
we can permute the entity parts, relation parts,
or flip signs
to get $4! \times 4! \times 2^4=9,216$ equivalents of $f^b$.
If 
%\footnote{$\surd$ what's this?}
$\mathcal G_{f^b}\cap\mathcal H^b\cap\mathcal T^b\neq\emptyset$,
we throw $f^b$ away since there are equivalent structures in the sampled set $\mathcal H^b$
and history record $\mathcal T^b$.
This step can dramatically help us to reduce the cost in training equivalent structures.
Take $f^4$ as an example, the whole space is reduced from $A_{16}^4\times 2^4$ to 5 through the filter,
namely there are only five good and unique candidates in $f^4$.
% picked by given in 
%\footnote{+++ Fig. 3 is duplicate with figure 2.
%	I think you'd better move it to appendix.}
%Fig.~\ref{fig:b4}.
Besides, we add exception for the condition in step~5  of Alg.\ref{alg:greedy} for $f^4$ 
since the number of candidates is smaller than $N$.

\subsubsection{Expressiveness - Constructing a predictor}
\label{ssec:pred}

%Since the performance of SFs on a specific KG is closely related to how the SF is formed,
%we can use a learning model,
%i.e.  the predictor $\mathcal P$,
%to predict the performance and avoid training on potentially poor SFs in advance.
%Considering that
%1) the performance is related to symmetric and asymmetric properties;
%and 2) collecting the real performance of each point in the search space is expensive,
%we are motivated to design symmetry-related features (SRFs),
%which can effectively capture to what extent $g(\bm{r})$ can be symmetric or skew-symmetric (Proposition~\ref{pr:srfs}),
%and has low complexity.
%The design principles and details are given in Appendix \ref{app:predictor}.
Even though the filter helps to throw away many unpromising candidates, 
it does not deal with Constraint~(C1).
Hence, after collecting $N$ candidates, 
we use the predictor $\mathcal P$ to further select $K_2$ promising ones among them.
Considering that the performance of SFs on a specific KG is closely related to how the SF is formed,
we can use a learning model, i.e. the predictor $\mathcal P$ 
to predict the performance and select good candidates in advance.
%The idea of using performance predictor is not new,
%it has been recently explored in algorithm selection \cite{eggensperger2015efficient},
%networks' performance prediction \cite{liu2018progressive}
%and hyper-parameter optimization \cite{feurer2015efficient}, etc.
In general,
we need to extract features for points 
which have been visited by the search algorithm,
and then use a learning model to predict validation performance based on those features \cite{feurer2015efficient,liu2018progressive}.
The following are principles
a good predictor needs to meet
\begin{itemize}[leftmargin=26px]
	\vspace{-1px}
	\item[(P1).] 
	\textit{Correlate well with true performance}:
	the predictor needs not to accurately predict the exact values of validation performance,
	instead it should rank good candidates over bad ones;
	
	\item[(P2).] 
	\textit{Learn from small samples}:
	as the real performance of each point in the search space is expensive to acquire,
	the complexity of the predictor should be low so that it can learn from few samples. 
	\vspace{-1px}
\end{itemize}

Based on Principle~(P1),
the extracted features from $g$ should be closely related to the quality of defined SF.
Meanwhile, the features should be cheap to construct,
i.e. they should not depend on values of $\bm{r}$,
which are unknown before training.
For (P2), the number of features should be small to guarantee a simple predictor.
Therefore, we are motivated to design the
symmetry-related features (SRFs),
which can effectively capture to what extent $g(\bm{r})$ can be symmetric or skew-symmetric (Proposition~\ref{pr:srfs}),
and has low complexity.

\begin{figure}[ht]
	\centering
	\vspace{-5px}
	\includegraphics[width=1.0\columnwidth]{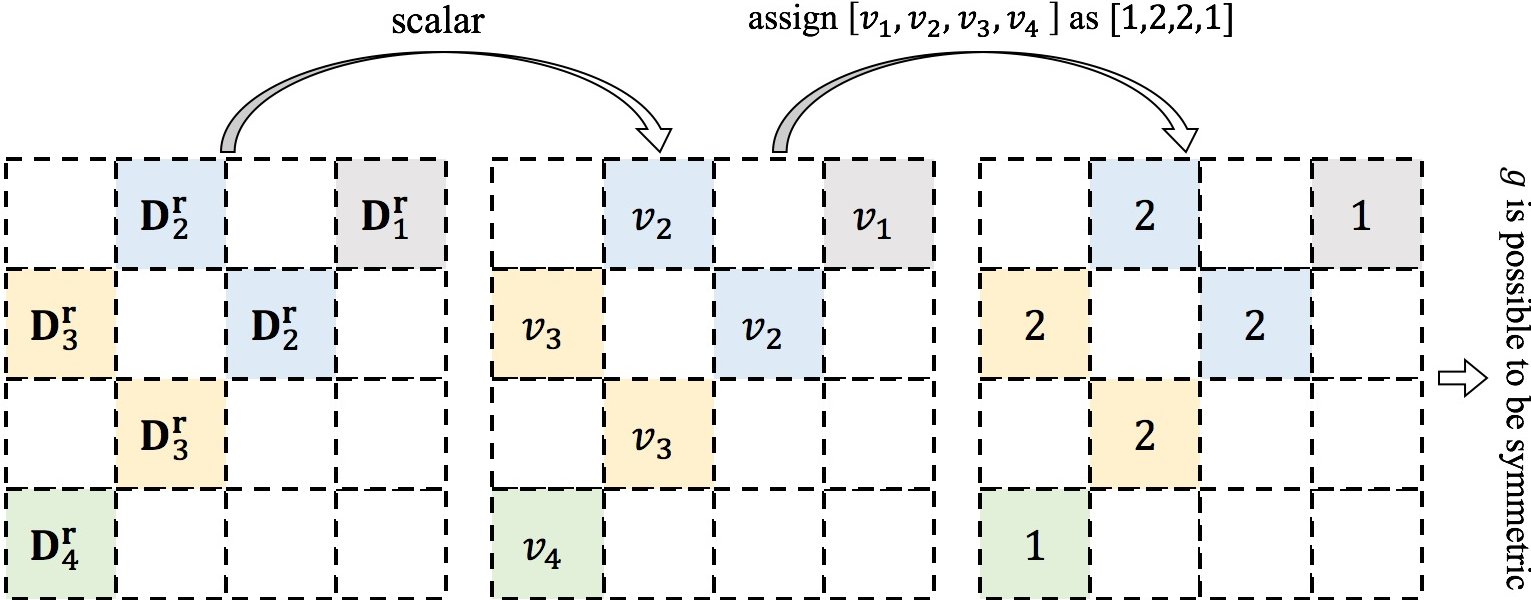}
	\vspace{-20px}
	\caption{Example of generating a feature of SRF.}
	\vspace{-10px}
	\label{fig:feat-gen}
\end{figure}

Similar as the filter, we also use a $4\times 4$ substitute matrix to represent $g$.
%\footnote{? adjust the size of the figure.}
As in Fig.~\ref{fig:feat-gen},
%\footnote{$\surd$ revise the caption,
%	what is S2?}
we use $\bm v=[v_1;v_2;v_3;v_4]$ to represent $[\bm{r}_1; \bm{r}_2; \bm{r}_3; \bm{r}_4]$,
then the symmetric and skew-symmetric property of $g$ can be checked through
$g(\bm v)-g(\bm v)^\top$ and $g(\bm v)+g(\bm v)^\top$.
Since $g(\bm v)$ is a simple $4\times 4$ matrix, the checking procedure is very cheap.
Then by assigning different values to $\bm v$ (details in Appendix~\ref{app:srf}),
a 22-dimensional SRF will be returned.
Considering that the correlation of SRFs with SFs' performance is guaranteed under Proposition~\ref{pr:srfs},
we can use a simple two-layer MLP (22-2-1) as the predictor $\mathcal P$.
Other regression models with low complexity may also work here.

\begin{prop} 
	\label{pr:srfs}
	The extracted SRFs (see Appendix~\ref{app:srf})
	are (i) invariant to both the permutations and flipping signs of blocks in $\bm{R}$ 
	and 
	(ii) give predictions related to symmetric or anti-symmetric properties.
\end{prop}

%\begin{remark}[Searching for better SFs]
%	It is possible that the optimal structure in Definition~\ref{def:unify}
%	is pruned when reducing the search space.
%	In this work, 
%	the goal is to search better SFs rather than the best one with as little cost as possible.
%	Besides, experiments show the efficiency and effectiveness of the search algorithms.
%\end{remark}

\subsection{Search complexity analysis}
\label{ssec:complexity}
%\vspace{-3px}

There are 16 blocks and each can be filled with 9 different contents 
$\{\mathbf 0$, $\pm\mathbf D^{\bm{r}}_1$, $\pm\mathbf D^{\bm{r}}_2$, $\pm\mathbf D^{\bm{r}}_3$, $ \pm\mathbf D^{\bm{r}}_4\}$.
Thus the whole space size is $9^{16}$, which is extremely large.
The greedy strategy,
predictor
and filter
cut down the space 
in different perspectives.
Specifically,
in each greedy step:
\begin{itemize}[leftmargin=10px,itemsep = 5px]

%Note that in
%each greedy step,
%there are only about $10^4$ possible candidates.
%Thus,
%the filter can 
%significantly reduce
%the number of candidates need to be trained.

\item \textit{Greedy}: 
%	\footnote{+++ this is our motivation to use greedy algorithm,
%		try to mention this at the start of this sub-section.}
Considering that $f^{b}$ is progressively generated on $f^{b-2}$ for $b=6,8,\dots$, 
there can be $C_{16-(b-2)}^2\times 4^2\times 2^2$ candidates
($C_{16-(b-2)}^2$ is to choose location, $4^2$ is for the two $\bm{r}_i$s and $2^2$ is for signs).
In comparison,
there can be $C_{16}^b\times 4^b \times 2^b$ possible SFs in $f^b$.
Take $b=6$ for example, 
there are  $2\times 10^9$ possible candidates.
Since $f^6$ is generated based on the 5 good candidates in $f^4$,
we reduce the space size from  $2\times10^9$ to approximately $3\times10^4$
based on the greedy scheme.
%Since top $K_1$ SFs in $\mathcal T^{b-2}$ are selected, 
%the per-step  search space for each $b>4$ is about $K_1\times 10^4$.
%Thus,
%the space induced 
%by the greedy strategy is $10^4 \times \nicefrac{(b - 4) K_1}{2}$,
%which is much smaller than $9^{16}$ from full space.
%In comparison, the space size of random search is $C_{16}^b\times (4\times 2)^b\gg K_1\times 10^4$.
%Thus, the greedy strategy can significantly cut down the search space.
%\vspace{-3px}

\item \textit{Filter}:
The filter we designed is mainly used to deal with invariance properties.
Permuting $\bm{r}_i$'s leads to $4!=24$ equivalent structures.
Simultaneously permuting $\bm{h}_i$'s
and $\bm{t}_i$'s also gives 24 equivalents.
Besides, there are $2^{4}=16$ possible signs patterns.
Therefore, given a $g(\bm{r})$, 
we can generate at most 
(there may exist same structures in this set) 
$24\times24\times16=9216$ equivalent SFs,
which should perform the same.
Besides, by constraining the SF 
under Constraint~(C2),
%to contain no zero/repeat rows/columns,
many bad candidates can also be filtered out.
Take $f^4$ as an example,
only 5 candidates are selected to be trained among approximately $700k$ possible structures.

%As for the Constraint~(C2), it is not easy to measure with certain numbers.
%But as can be seen in Fig.~\ref{fig:b4}, 
%the filter reduces the whole space size from $A_{16}^4 \times 2^4$ to 5,
%a very significant reduction.

\item \textit{Predictor}:
Once $N$ candidates are generated, 
the predictor will select $K_2$ ones based on their predicted performance.
Thus, the reducing ratio of predictor is about ${N}/{K_2}$.

\end{itemize}

While 
it is difficult to 
directly quantize
to which extend
the three steps together 
can help to reduce the search space,
we can observe the significance of efficiency gained through each component.
Besides, 
we perform an empirical study
in Sec.\ref{ssec:ablation}
to show the performance gaining of these steps.

\begin{table*}[ht]
	\centering
	\caption{Statistics of the data sets used in experiments. ``sym'' and ``anti-sym'' denotes the symmetric and anti-symmetric relations.}
	\label{tab:dataset}
	\vspace{-9px}
	\begin{tabular}{c|ccccc|cccc}
		\hline
		                data set                 & \#entity & \#relation &  \#train  & \#valid & \#test & \#sym & \#anti-sym & \#inverse & \#general \\ \hline
		   WN18 \cite{bordes2013translating}     &  40,943  &     18     &  141,442  &  5,000  & 5,000  &   4   &     7      &     7     &     0     \\
		  FB15k  \cite{bordes2013translating}    &  14,951  &   1,345    &  484,142  & 50,000  & 59,071 &  66   &     38     &    556    &    685    \\
		WN18RR  \cite{dettmers2017convolutional} &  40,943  &     11     &  86,835   &  3,034  & 3,134  &   4   &     3      &     1     &     3     \\
		 FB15k237 \cite{toutanova2015observed}   &  14,541  &    237     &  272,115  & 17,535  & 20,466 &  33   &     5      &    20     &    179    \\
		 YAGO3-10 \cite{mahdisoltani2013yago3}   & 123,188  &     37     & 1,079,040 &  5,000  & 5,000  &   8   &     0      &     1     &    28     \\ \hline
	\end{tabular}
	\vspace{-5px}
\end{table*}

\subsection{Comparison with existing AutoML approaches}
The most related work in the AutoML literature is PNAS \cite{liu2018progressive},
which combines a greedy algorithm with a performance predictor to search 
a cell structure for the convolutional neural network (CNN).
However,
the filter is not used in PNAS 
as the search space for AutoSF is fundamentally different from that of CNN.
Besides,
PNAS adopts direct one-hot encoding for the predictor,
which has a bad empirical performance here (see Sec.\ref{sub:fpref}).
%In detail, we can encode each $\left[ g(\bm{R}) \right]_{ij}$
%$\in \{ \mathbf{0}$, 
%$\pm \bm{R}_1$, 
%$\pm \bm{R}_2$, 
%$\pm \bm{R}_3$, 
%$\pm \bm{R}_4 \}$
%into a one-hot vector
%and concatenate the vectors for the 16 $\left[ g(\bm{R}) \right]_{ij}$'s with $i,j=1\dots 4$.
%For example,
%if $\left[ g(\bm{R}) \right]_{ij} = - \bm{R}_1$,
%such block is then encoded as two parts: $\left[ 1, 0, 0, 0\right]$ for indexing $\bm{r}_1$ 
%and $[0,1]$ for the sign $-1$. 
%Besides, $\mathbf 0$ will be represented as a 6-dim zero vector.
%As a result, a 96-dimensional feature vector will be extracted.
%Instead,
%th
%When comparing with the 22-dimensional SRFs,
%the one-hot representation lacks guarantees for capturing expressiveness and invariance properties,
%and has larger complexity.
As for the other AutoML approaches, 
even though the search problem of AutoSF is similarly defined as HPO \cite{bergstra2011algorithms,eggensperger2015efficient}
and NAS \cite{elsken2019neural},
the search space and search algorithm of AutoSF
are novel and specifically designed for KGE.
There is no direct way for them to deal with the challenges
in Sec.\ref{ssec:challange}.
%More recently, 
%one-shot architecture search methods like DARTS \cite{liu2018darts}
%become popular in NAS area due to its efficiency by sharing parameters.
%But for AutoSF,
%different models cannot be fairly compared  when embeddings are shared,
%which means one-shot architecture search methods cannot be adopted in our problem.

\begin{table*}[ht]
	\caption{Comparison of the best SF identified by AutoSF and the state-of-the-art SFs. 
		The bold number means the best performance, and the underline  means the second best. 
		DistMult, ComplEx, Analogy and SimplE are obtained from our implementation,
		others are copied from the corresponding reference paper.
		STD is less than 0.001, thus not reported.}
	\vspace{-9px}
	\label{tb:comparison}
	\centering
	\begin{tabular}{c|c|ccc|ccc|ccc|ccc|ccc}
		\hline
		&                                         &      \multicolumn{3}{c|}{WN18}       &     \multicolumn{3}{c|}{FB15k}      &     \multicolumn{3}{c|}{WN18RR}      &    \multicolumn{3}{c|}{FB15k237}     &     \multicolumn{3}{c}{YAGO3-10}     \\ \cline{3-17}
		type &                  model                  &        MRR      & $\!\!$H@1$\!\!$  &       $\!\!$H@10 $\!\!$      &       MRR       & $\!\!$H@1$\!\!$  &       $\!\!$H@10$\!\!$       &        MRR        & $\!\!$H@1$\!\!$ &       $\!\!$H@10$\!\!$       &        MRR       & $\!\!$H@1$\!\!$  &       $\!\!$H@10$\!\!$       &        MRR       & $\!\!$H@1$\!\!$  &       $\!\!$H@10$\!\!$       \\ \hline
		$\!\!\!\!$TDM$\!\!\!\!$  &    TransE \cite{zhang2018nscaching}     &       0.500  &   ---      &       94.1       &      0.495     &      ---    &       77.4       &       0.178    &   ---        &       45.1     &    0.256       &  ---      &       41.9       &         ---              &   ---     &        ---         \\
		&    TransH \cite{zhang2018nscaching}     &       0.521     & ---   &       94.5       &      0.452    &      ---     &       76.6       &       0.186     &    ---      &       45.1       &       0.233     &      ---    &       40.1       &         ---        &      ---   &        ---         \\
		&                 RotatE \cite{sun2019rotate}                 &       0.949     & 94.4   &       95.9       &      0.797      &  74.6  &       88.4       &       \underline{0.476}    &    42.8   &       \textbf{57.1}       &       0.338  &  24.1   &       53.3       &        ---      &     ---     &       ---        \\ \hline\hline
		$\!\!\!\!$NNM$\!\!\!\!$  &      NTN \cite{yang2014embedding}       &       0.53       &     ---    &       66.1       &       0.25       &        &       41.4       &         ---       &     ---     &        ---         &     ---           &    ---      &        ---           &         ---        &     ---        &        ---         \\
		& $\!\!$Neural LP \cite{yang2017differentiable}$\!\!$ &       0.94         &   ---   &       94.5       &       0.76        &  ---  &       83.7       &         ---         &      ---      &        ---         &       0.24         &   ---  &       36.2       &         ---        &      ---       &        ---         \\
		& ConvE \cite{dettmers2017convolutional}  &       0.942         &  93.5   &      {95.5}      &      0.745          &      67.0   &       87.3       &       0.46        &  39.    &       {48.}       &      {0.316}       & 23.9    &       49.1       &       0.52        &       45.     &      {66.}      \\ \hline\hline
		$\!\!\!\!$BLM$\!\!\!\!$  &   TuckER \cite{balavzevic2019tucker}    &  \textbf{0.953}     & \textbf{94.9}    &       95.8       &      0.795       &   74.1   &       89.2       &       0.470        &  \underline{44.3}     &       52.6       & \underline{0.358}    &   \underline{26.6}   &       54.4       &         ---           &    ---      &        ---         \\
		&      HolEX \cite{xue2018expanding}      &       0.938      &    93.0   &       94.9       &      0.800        &    75.0   &       88.6       &         ---         &     ---       &        ---         &         ---         &     ---       &        ---         &         ---          &    ---       &        ---         \\
		&	QuatE  \cite{zhang2019quaternion}        &   0.950    &  94.5  & 95.9   & 0.782  &  71.1  &  90.0  &  \underline{0.488}  &  43.8  &  \textbf{58.2} &   0.348  &  24.8 &  55.0  &  ---  &  ---  &   ---  \\ 
		&                DistMult                 &       0.821      &      71.7       &       95.2       &      0.817      &        77.7     &       89.5       &       0.443         &  40.4   &       50.7       &      {0.349}       &  25.7  &       53.7       &       0.552          &   47.6    &       69.4       \\
		&                 ComplEx                 &       0.951          &     94.5      &       95.7       &      \underline{0.831}       &    79.6   &  \underline{90.5}   &      {0.471}        &   43.0 	 &       55.1       &       0.347         &     25.4		 &       54.1       & \underline{0.566}      &    \underline{49.1}   &       70.9       \\
		&                 Analogy                 &       0.950         &    94.6   &       95.7       &      0.829      &   79.3   &  \underline{90.5}   & {0.472}    &    43.3   & {55.8} &       0.348       &  25.6    & \underline{54.7} &       0.565         &   49.0    & \underline{71.3} \\
		&                SimplE/CP                &       0.950        &   94.5   & \underline{95.9} &      0.830        &   \underline{79.8}    &      {90.3}      &      {0.468}         &   42.9    &       55.2       &      {0.350}    &    26.0    &       54.4       &       0.565      &   \underline{49.1}   & {71.0} \\ \hline\hline
		\multicolumn{2}{c|}{AnyBURL \cite{meilicke2019anytime}}       &  0.95 & 94.6 &  \underline{95.9}  & 0.83   &  80.8 &  87.6  &  0.48  & 44.6   &  55.5   &  0.31   &  23.3 &    48.6       &   0.54  &  47.7   &  47.3        \\ \hline \hline
		\multicolumn{2}{c|}{AutoSF}          & \underline{0.952}    &   \underline{94.7}    &  \textbf{96.1}   &  \textbf{0.853}     &  \textbf{82.1}   & \textbf{91.0} &  \textbf{0.490}      &   \textbf{45.1}   &  \underline{56.7}   &  \textbf{0.360}    &    \textbf{26.7}  &  \textbf{55.2}   &  \textbf{0.571}     &  \textbf{50.1}  &  \textbf{71.5}   \\ \hline
	\end{tabular}
	\vspace{-13px}
\end{table*}

\section{Empirical Study}

%In this section,
%we first set up experiments in Sec.\ref{ssec:setup}.
%Then,
%in Sec.\ref{ssec:compstat} and \ref{ssec:tripclass}, 
%we show AutoSF's ability in finding SFs of which the performance can beat human-designed ones.
%The efficiency of AutoSF is presented in Sec.\ref{ssec:comprand} by comparing with other search schemes.
%We do various ablation study in Sec.\ref{ssec:ablation} to analyze the search algorithm in steps.
All of the algorithms are written in python with PyTorch framework \cite{paszke2017automatic}.
Experiments are run
on 8 TITAN Xp GPUs.

\subsection{Experiment setup}
\label{ssec:setup}
\subsubsection{Datasets}
Five data sets, i.e.  WN18, FB15k, WN18RR, FB15k237 and YAGO3-10 are considered
(statistics in Tab.~\ref{tab:dataset}).
WN18RR and FB15k237 are variants 
that remove near-duplicate or inverse-duplicate relations from WN18 and FB15k respectively,
\cite{wang2018evaluating,toutanova2015observed}.
YAGO3-10 is much larger than the others.
These are benchmark datasets,
and are popularly used to compare KGE models in the literature
\cite{bordes2013translating,yang2014embedding,trouillon2017knowledge,liu2017analogical,kazemi2018simple,lacroix2018canonical}.
%Besides, we roughly compute the number of symmetric, anti-symmetric and inverse pairs of each data sets.
%Statistics of the data sets we used are listed in Tab.~\ref{tab:dataset}.
%\footnote{$\surd$ you have claimed 4 types in Tab.~\ref{tab:comrel}.
%	Thus it is also better to have 4 columns here}
The number of symmetric, anti-symmetric, inverse pairs and general asymmetric are computed in the following way:
Given a relation $r$, let the number of positive triplets $(h,r,t)$ be $n_r$.
(i) If the number of $(t,r,h)$ is larger than $0.9n_r$, then we regard it as symmetric;
(ii) If the number of $(t,r,h)$ is zero and the size of joint set of $h$ and $t$ is at least $0.1n_r$
(this is to ensure that they have same type),
we regard as anti-symmetric;
(iii) If there exist another relation $r'$ that has at least $0.9n_r$ $(t,r',h)$,
then $r$ and $r'$ are inverse pairs;
(iv) others are regarded as general asymmetric.
The threshold 0.9 and 0.1 are hand-made and just used to roughly (other values are fine) indicate the relation properties for each data set.

\subsubsection{Hyper-parameters}
\label{ssec:hyper:setting}
%\footnote{$\surd$ this part needs revise.}
%Based on 
%\footnote{$\surd$ yeah, what is ``some''?}
%some preliminary experiments,
%we find that the overall performance is sensitive to hyper-parameters
%but the relative ranking of different SFs is stable for fixed set of hyper-parameters.
Since the searched embedding models belong to BLMs,
we fairly compare different SFs with a fixed set of hyper-parameters.
In order to reduce training time,
we set the dimension $d$ as 64 during the search procedure.
First,
we use SimplE \cite{kazemi2018simple} as the benchmark model 
and tune hyper-parameters with the help of HyperOpt,
a hyper-parameter optimization framework based on TPE \cite{bergstra2011algorithms}. 
The searching ranges are given as follows: 
learning rate $\eta$ in $[0, 1]$, 
L2 penalty $\lambda$ in $[10^{-5}, 10^{-1}]$,
decay rate in $[0.99, 1.0]$,
batch size $m$ in $\{256, 512, 1024\}$.
All the models are trained until converge
to avoid the influence of different convergence speed.
Besides, we use Adagrad \cite{duchi2011adaptive} as the optimizer
since it tends to perform better as indicated in \cite{lacroix2018canonical,trouillon2017knowledge}.
Once a good hyper-parameter configuration is selected,
we use it to train and evaluate different searched SFs.
After the search procedure, we pick up the best SF evaluated by the MRR performance 
on the validation data set
as the searched SF.
When comparing the searched SFs with human-designed ones,
we increase the dimension from 64 to $d\in \{256, 512, 1024, 2048\}$ as in \cite{lacroix2018canonical}.
As mentioned in \cite{wang2018evaluating},
KGE models are sensitive to hyper-parameters.
For a fair comparison,
we use the same set of  hyper-parameters to train and evaluated different models on each dataset.

\subsubsection{Meta Hyper-parameters}
The hyper-parameters
$K_1, K_2$ and $N$ have little influence to the search procedure. 
We use $K_1=K_2=8$ and $N=256$ for all data sets.
Besides, steps~\ref{step:gen-start}-\ref{step:update} in Alg.\ref{alg:greedy} run based on an inner loop. 
We train 8 models in parallel and iterate for 32 times (16 times for YAGO3-10),
namely we evaluate 256 $f^b$s for each $b>4$.

\begin{figure*}[ht]
	\centering
	\subfigure[WN18.]
	{\includegraphics[height=2.7cm]{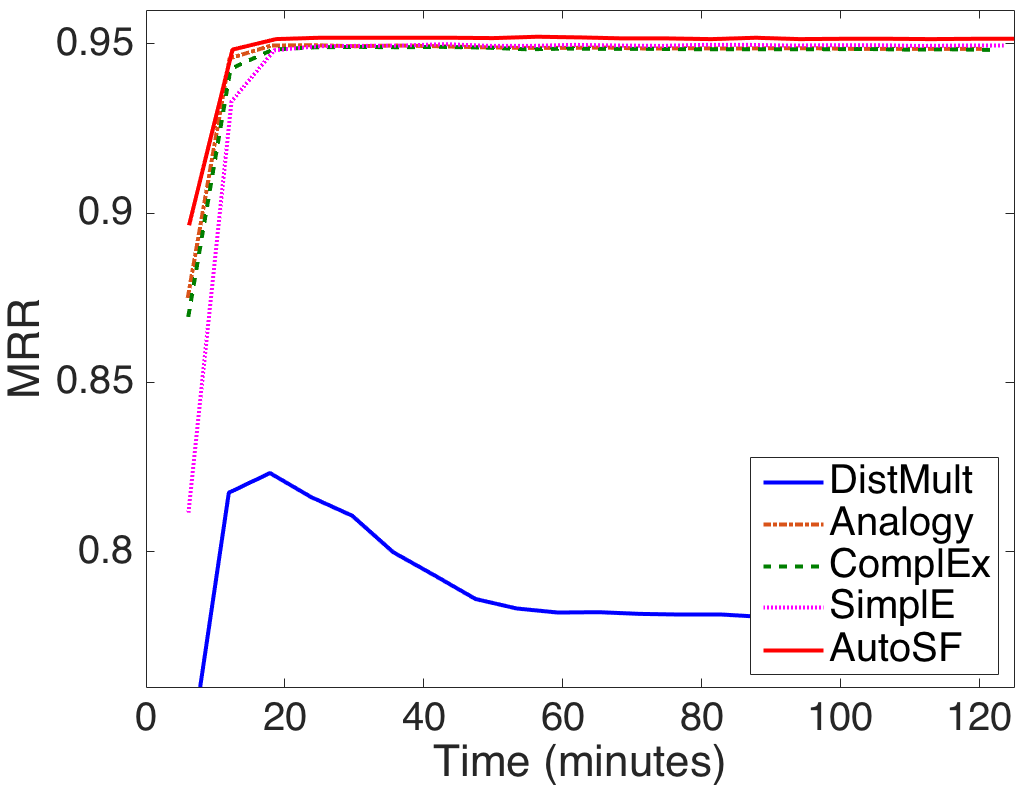}}
	\subfigure[FB15k.]
	{\includegraphics[height=2.7cm]{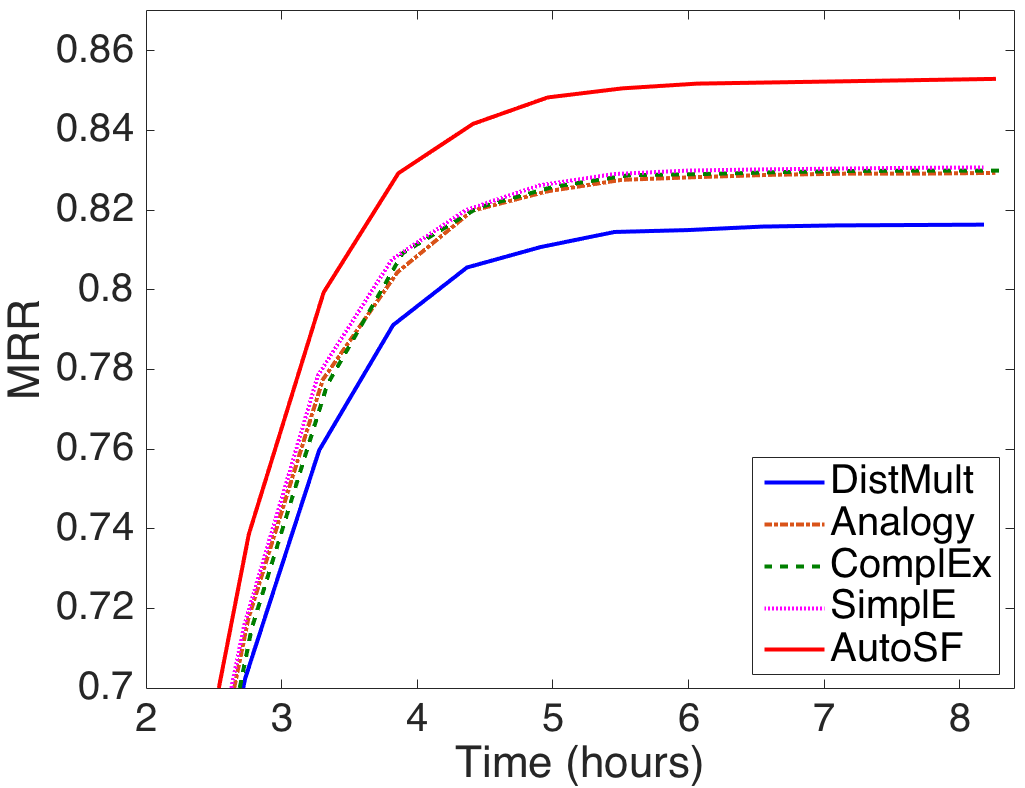}}
	\subfigure[WN18RR.]
	{\includegraphics[height=2.7cm]{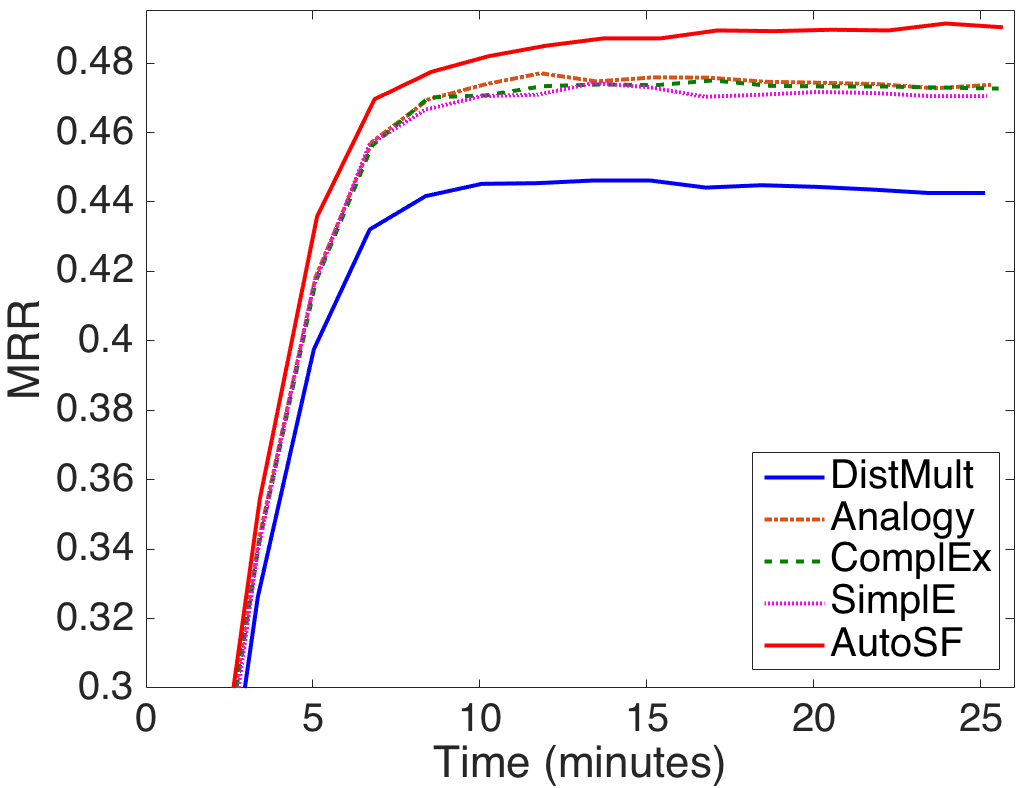}}
	\subfigure[FB15k237.]
	{\includegraphics[height=2.7cm]{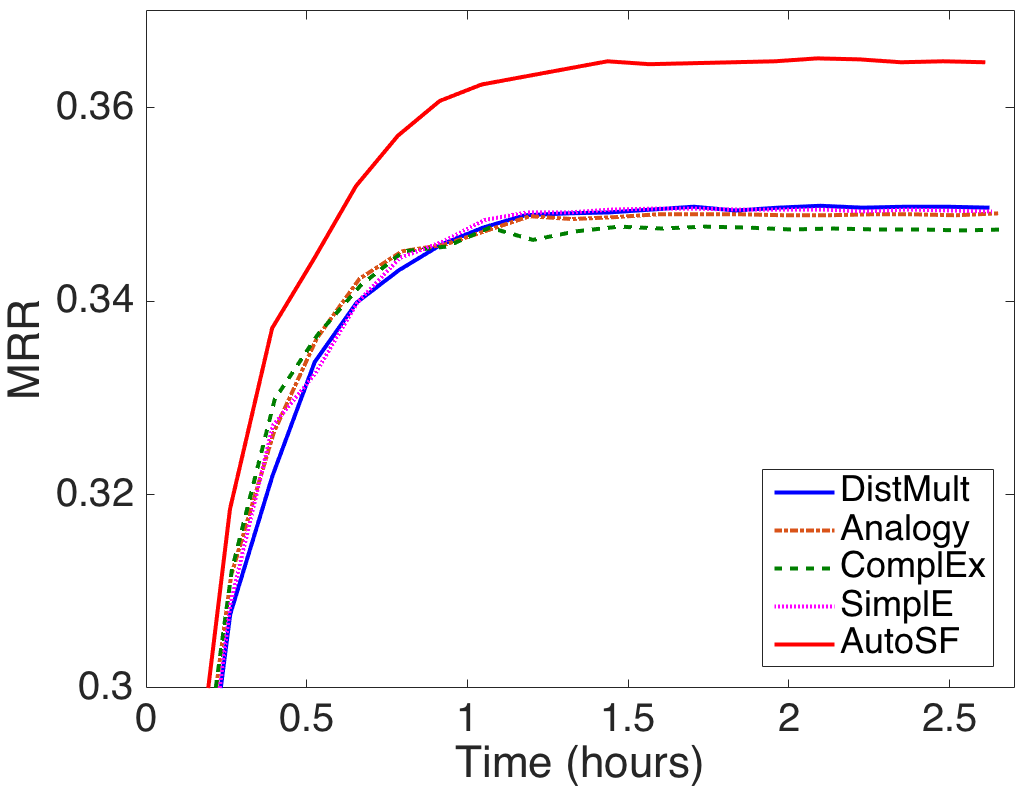}}
	\subfigure[YAGO3-10.]
	{\includegraphics[height=2.7cm]{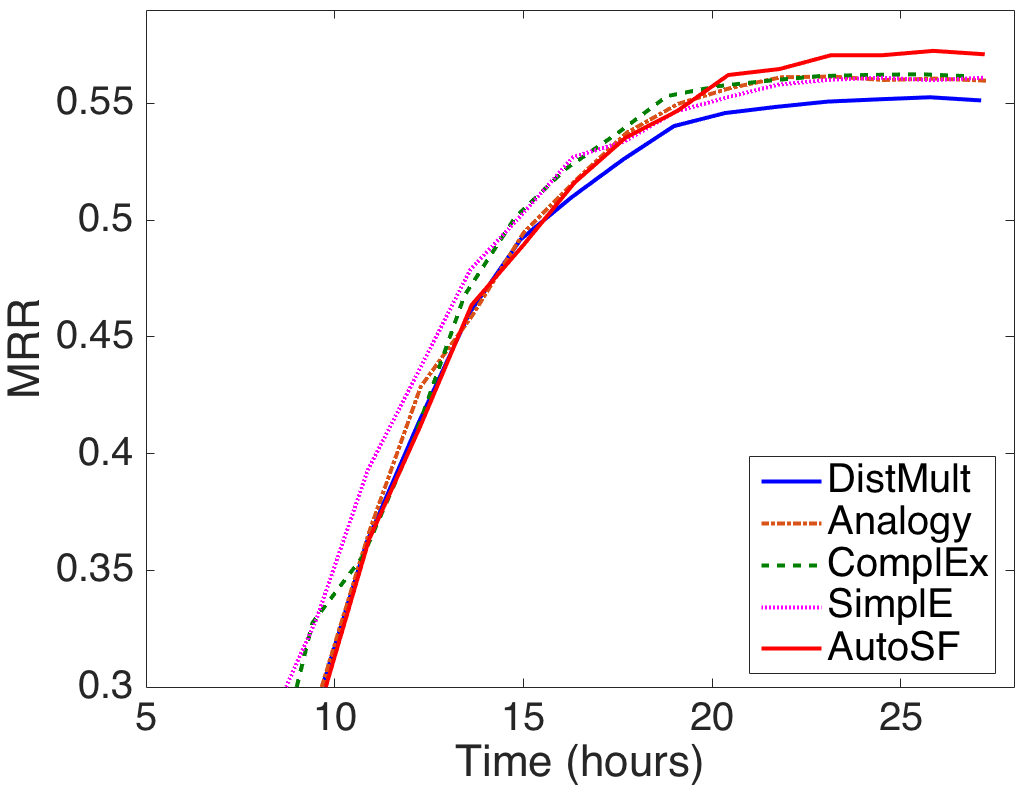}}
	\vspace{-6px}
	\caption{Comparison on clock time (in hours) of model training
		v.s testing MRR 
		between search SFs (by AutoSF) and 
		human-designed ones.}
	\label{fig:curve}
		\vspace{-5px}
\end{figure*}

\begin{figure*}[ht]
	\centering
	\subfigure[WN18.]
	{\includegraphics[width=0.32\columnwidth]{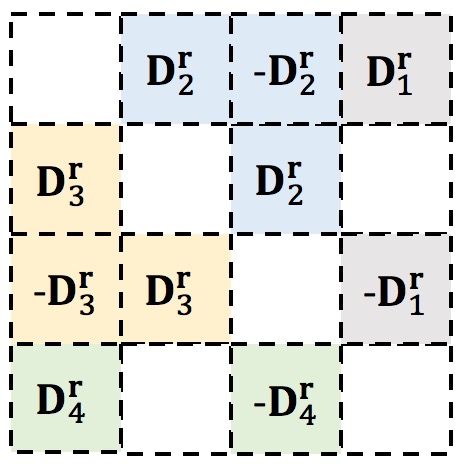}}
	\quad
	\subfigure[FB15k.]
	{\includegraphics[width=0.32\columnwidth]{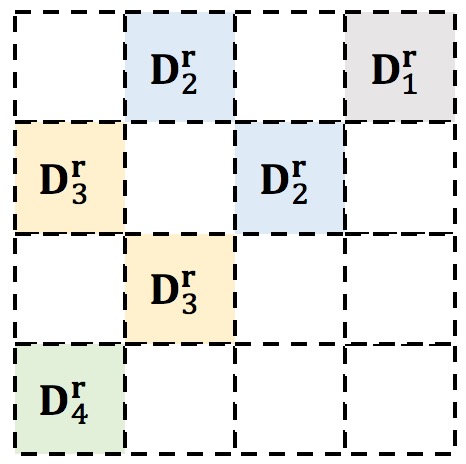}}
	\quad
	\subfigure[WN18RR.]
	{\includegraphics[width=0.32\columnwidth]{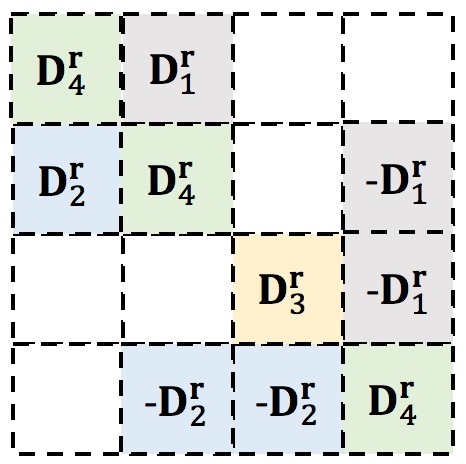}}
	\quad
	\subfigure[FB15k237.]
	{\includegraphics[width=0.32\columnwidth]{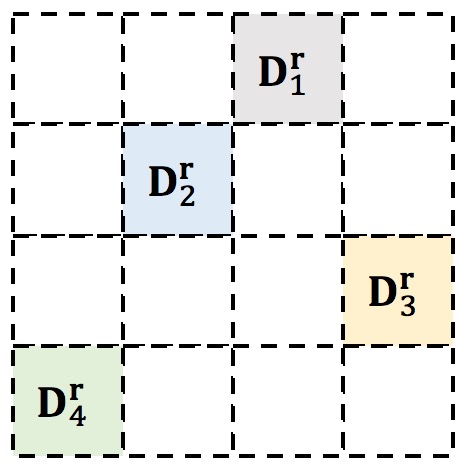}}
	\quad
	\subfigure[YAGO3-10.]
	{\includegraphics[width=0.32\columnwidth]{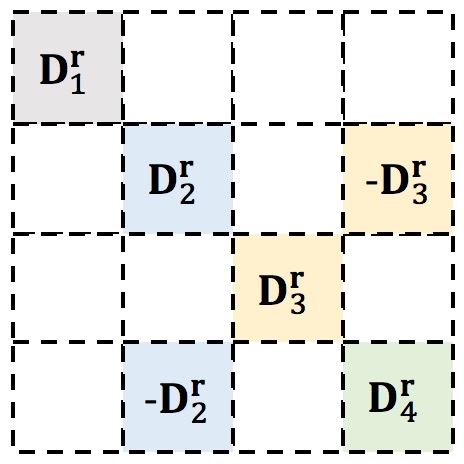}}
	\vspace{-6px}
	\caption{A graphical illustration of SFs identified by our AutoSF on each data set.}
	\label{fig:searchedsf}
	\vspace{-12px}
\end{figure*}

\subsection{Comparison with existing SFs on link prediction}
\label{ssec:compstat}
%\vspace{-5px}

We compare our AutoSF with the state-of-the-art KGE models
discussed in Sec.\ref{ssec:score},
which are designed by humans,
i.e. 
TransE \cite{bordes2013translating},
TransH \cite{wang2014knowledge},
and
RotatE \cite{sun2019rotate}
from TDMs;
NTM \cite{socher2013reasoning},
Neural LP \cite{yang2017differentiable},
and 
ConvE \cite{dettmers2017convolutional}
from NNMs;
{
TuckER \cite{balavzevic2019tucker},
HolE/HolEX \cite{nickel2016holographic,xue2018expanding}, 
Quat \cite{zhang2019quaternion},
DistMult \cite{yang2014embedding},
ComplEx \cite{trouillon2017knowledge},
Analogy \cite{liu2017analogical}
and
SimplE \cite{kazemi2018simple}
from BLMs;
and a rule-based method AnyBURL \cite{meilicke2019anytime}.
}
Hyper-parameters are selected by the MRR value on the validation set.
%\footnote{$\surd$ we no longer have this}
%(see Appendix~\ref{ssec:hyperdetails}).

Following \cite{yang2014embedding,trouillon2017knowledge,liu2017analogical,kazemi2018simple,dettmers2017convolutional},
we test KGE's performance based on 
\textit{link prediction}.
For each triplet $(h,r,t) \in \mathcal{S}$, where $\mathcal{S}$ is the validation or testing set,
we compute the score of $(h',r,t)$ for all $h'\in\mathcal E$ and get the rank of $h$,
the same for $t$ based on scores of $(h,r,t')$ over all $t'\in\mathcal E$,
$r$ is not compared as in the literature \cite{wang2017knowledge}.
Same as above mentioned papers, 
we adopt the following metrics: 
(i) 
Mean reciprocal ranking (MRR): 
%It is computed by average of the reciprocal ranks 
{\small $\nicefrac{1}{|\mathcal{S}|}\sum_{i=1}^{|\mathcal{S}|}\nicefrac{1}{\text{rank}_i}$},
where $\text{rank}_i, i\in\{1,\dots,|\mathcal{S}|\}$ is a set of ranking results
and
(ii) H@10: 
%It is the percentage of appearance in top-$10$ ranking: 
{\small $\nicefrac{1}{|\mathcal{S}|}\sum_{i=1}^{|\mathcal{S}|}\mathbb I\left(\text{rank}_i<10\right)$}, 
where $\mathbb I(\cdot)$ is the indicator function.
We report the performance in a ``filtered'' setting as in \cite{bordes2013translating,wang2014knowledge},
where larger MRR and H@10 indicate higher embedding quality.

\subsubsection{Effectiveness}
A comparison of the testing performance
of AutoSF and 
the current state-of-the-art SFs
are shown in 
Tab.~\ref{tb:comparison}.
Firstly, we can see that there is no absolute winner among the baseline SFs.
For example,
TuckER is the best on WN18,
but is the worst among human-designed BLMs on FB15k.
DistMult generally performs worse on the benchmarks except for FB15k237
since it does not follow Proposition~\ref{pr:expre}.
A single model is hard to adapt to different KGs.
However,
AutoSF
performs consistently well among
these five data sets.
i.e. 
the best among FB15k, WN18RR, FB15k237 and YAGO3-10,
and the runner-up on WN18.

Besides, we plot the learning curves of DistMult, Analogy, ComplEx, SimplE and the best SF
searched by AutoSF in Fig.~\ref{fig:curve}.
As shown, the searched SFs not only outperform baselines, but also converge faster,
which may due to these SFs can better capture relations in these datasets.
%An interesting finding is that Analogy, ComplEx and SimplE share very similar learning curves
%since they have similar patterns in modeling symmetric and anti-symmetric relations.
%%The demonstrate that the SFs proposed in literature become 
%%\footnote{$\surd$ no ``marginal'' in this version}
%%marginal.
%Therefore, 
%it is meaningful for us to explore better SFs with the power of AutoML.

\subsubsection{Case study: Distinctiveness}
\label{ssec:casestufy}

To show the searched SFs are KG-dependent and novel to the literature,
we plot them in Fig.~\ref{fig:searchedsf}.
It is obvious that these SFs are different from each other,
and they are not equivalent regarding invariance properties.
As shown in Tab.~\ref{tab:dataset}, WN18 and FB15k have many symmetric, anti-symmetric relations
and inverse relation pairs, the best SF searched on them are very similar
and have the same SRF.
The other three data sets are more realistic
and contains less symmetric, anti-symmetric and inverse relations,
thus have different SRFs with fewer entry being non-zero.

The most special case is FB15k237, which can only be symmetric under (S11).
Viewing the values in Tab.~\ref{tb:comparison},
we can see that the leading performance on FB15k237 is achieved by DistMult and AutoSF,
both of which cannot be skew-symmetric.
As given in the statistic information in Tab.~\ref{tab:dataset},
FB15k237 has relatively fewer anti-symmetric relations.
This may explain why skew-symmetric is not that important for $g(\bm{r})$.
%This may explain why such a $g(\bm{r})$ can outperform the others.
%In addition, as indicated in \cite{dettmers2017convolutional},
%FB15k237 has more indegree nodes which cares more about the difference between multiple heads, 
%like the relation \textit{was\_born\_in}, rather than the direct plausibility of triplets.
%The information of in-degree and out-degree will be considered as additional information in our future work.
However,
SRFs still work for these cases
since it can be aware that skew-symmetric property is not that essential
and focus more on searching different local structures.

\begin{table}[ht] 
\centering
\vspace{-10px}
\caption{MRRs of applying SF searched from one data set (indicated by each row) on another data set (indicated by each column).}
\vspace{-10px}
\label{tab:dist}
\begin{tabular}{c|c|c|c|c|c}
	\hline
	                     &      WN18      &     FB15k      & $\!\!$WN18RR$\!\!$ & $\!\!$FB15k237$\!\!$ & $\!\!$YAGO3-10$\!\!$ \\ \hline
	        WN18         & \textbf{0.952} &     0.841      &       0.473        &        0.349         &        0.561         \\ \hline
	       FB15k         &     0.950      & \textbf{0.853} &       0.470        &        0.350         &        0.563         \\ \hline
	       WN18RR        &     0.951      &     0.833      &   \textbf{0.490}   &        0.345         &        0.568         \\ \hline
	$\!\!$FB15k237$\!\!$ &     0.894      &     0.781      &       0.462        &    \textbf{0.360}    &        0.565         \\ \hline
	$\!\!$YAGO3-10$\!\!$ &     0.885      &     0.835      &       0.466        &        0.352         &    \textbf{0.571}    \\ \hline
\end{tabular}
\vspace{-5px}
\end{table}

Besides,
we pick up the best SF searched from one data set
and test it on another data set
in Tab.~\ref{tab:dist}.
%\footnote{+++ visualization different SFs are OK, move this table to appendix.}
We can readily
find that these SFs get the best performance
on the data sets where they are searched.
This again demonstrate that
SFs found by AutoSF on different KGs are distinct from each other.

\subsection{Comparison with existing SFs on triplet classification}
\label{ssec:tripclass}

To further demonstrate the effectiveness of the searched SFs,
we do triplet classification as in \cite{wang2014knowledge}.
This task is to confirm whether a given $(h,r,t)$ is correct or not
and is more helpful in answering yes-or-no questions.
The decision rule of classification is as follows:
for each $(h,r,t)$,
if its score is larger than the relation-specific threshold $\sigma_r$,
which we predict to be positive,
otherwise negative.
The threshold $\sigma_r$ is determined by maximizing the accuracy 
of the validation set.
We test this task on FB15k, WN18RR and FB15k237, in which the positive and negative triplets are provided.
As shown in 
%\footnote{+++ hows STD?}
Tab.~\ref{tab:trip:class}, searched SFs consistently outperform human-designed BLMs.

\begin{table}[H]
	%	\end{minipage}
	%	\hfill
	%	\begin{minipage}[b]{0.41\linewidth}
	\centering
	\vspace{-5px}
	\caption{Comparison of searched SFs with the state-of-the-art SFs on accuracy (in \%) for triplet classification. STD$<$0.2.}
	\label{tab:trip:class}
	\vspace{-10px}
	\begin{tabular}{c|ccc}
		%	\vline
		\hline &     FB15k     &    WN18RR     &   FB15k237    \\ \hline
		DistMult      &     80.8      &     84.6      &     79.8      \\ \hline
		Analogy       &     82.1      &     86.1      &     79.7      \\ \hline
		ComplEx       &     81.8      &     86.6      &     79.6      \\ \hline
		SimplE       &     81.5      &     85.7      &     79.6      \\ \hline
		AutoSF       & \textbf{82.7} & \textbf{87.7} & \textbf{81.2} \\ \hline
	\end{tabular}
	\vspace{-5px}
\end{table}

\subsection{Comparison with other AutoML approaches} 
\label{ssec:comprand}
%\vspace{-5px}

\begin{figure}[ht]
	\centering
	\vspace{-10px}
	\includegraphics[height=3cm]{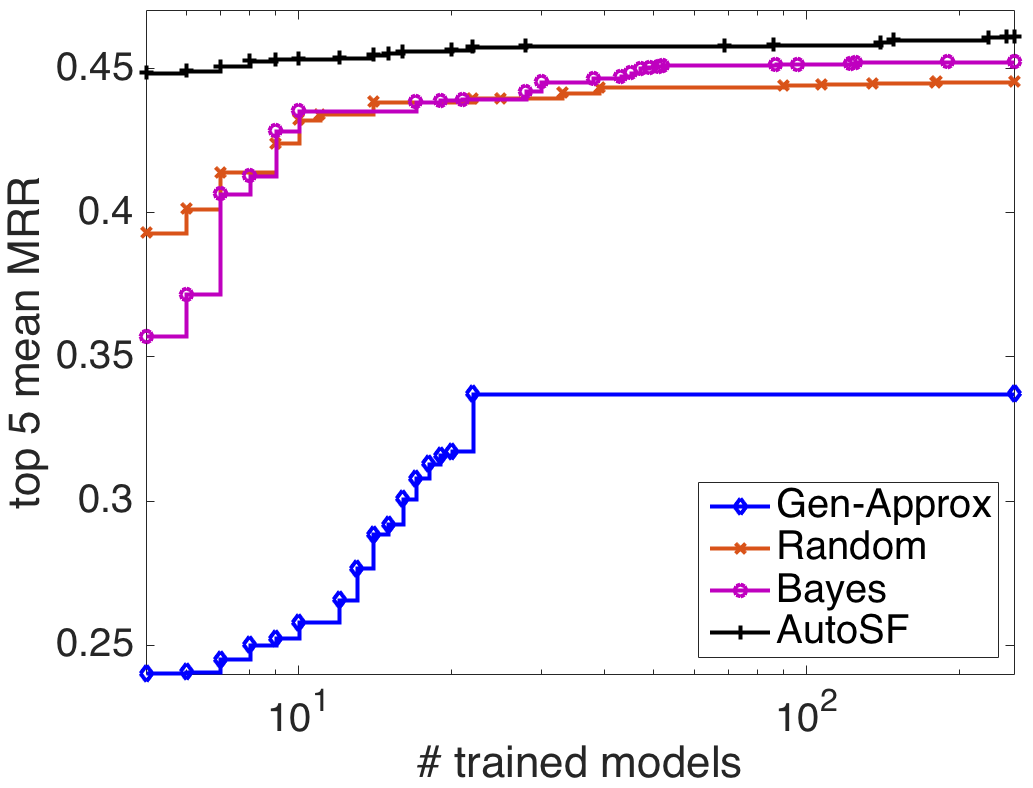}
	\quad
	\includegraphics[height=3cm]{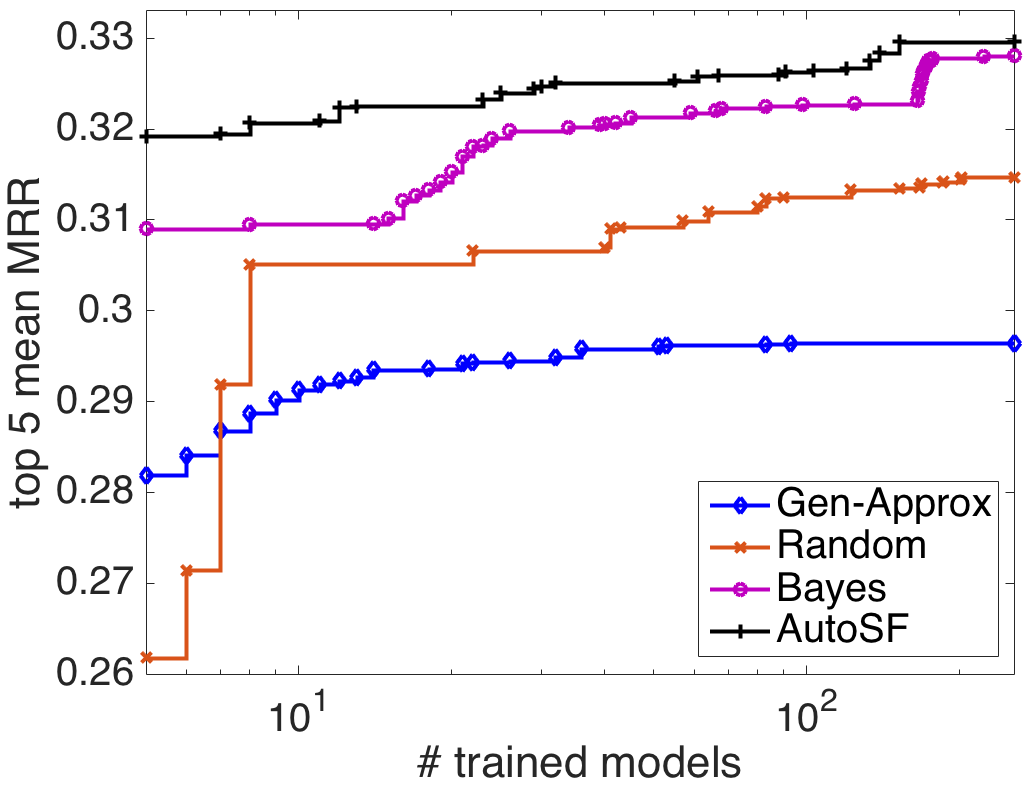}
	\vspace{-8px}
	\caption{Comparison of AutoSF with other AutoML approaches.}
	\label{fig:automl}
	\vspace{-15px}
\end{figure}

%\subsubsection{Effectiveness of the Search Space}
In this part, we compare AutoSF with the other search algorithms.
WN18RR and FB15k237 are used here, and all algorithms share the same set of hyper-parameters.
First, to show the effectiveness of the search space in BLM, 
we train a general approximator (\textit{Gen-Approx}), i.e.  MLP (in Appendix~\ref{app:mlp}), on the validation set.
Then, AutoSF is compared with \textit{Random} search and \textit{Bayes} algorithm 
\cite{bergstra2011algorithms}
on $f^6$. 
As shown in Fig.~\ref{fig:automl},
the general approximator performs much worse than BLM
since it is too flexible to consider domain-specific constraints and easily overfits.
For BLM settings, the Bayes algorithm can improve the efficiency upon random search.
However, it will easily fall into local optimum
and does not take the domain property into account.
Among them,
AutoSF is the most efficient and has the best any-time performance.
%by cutting down the search space.
%Note that,
%as discussed in Sec.\ref{ssec:challange},
%other popular search algorithms,
%e.g.,
%reinforcement learning \cite{zoph2017neural,baker2017designing},
%Bayes optimization \cite{feurer2015efficient}
%and genetic programming \cite{xie2017genetic} are not applicable and thus not compared. 

\subsection{Ablation study}
\label{ssec:ablation}
We use WN18RR and FB15k237 to illustrate the importance of different components
in the proposed searching algorithm.
%\footnote{$\surd$ which data sets are used here.}

\subsubsection{Filter and predictor}
\label{sub:fpref} 
\begin{figure}[ht]
	\centering
	\vspace{-8px}
	\includegraphics[height=3cm]{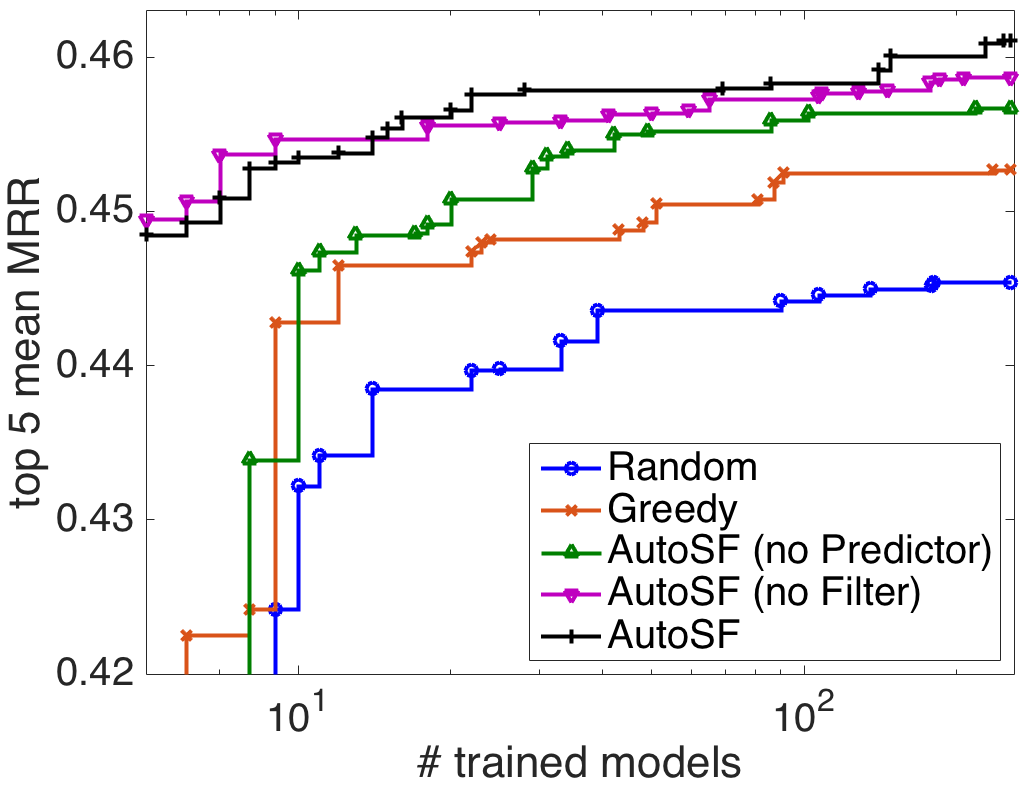}
	\quad
	\includegraphics[height=3cm]{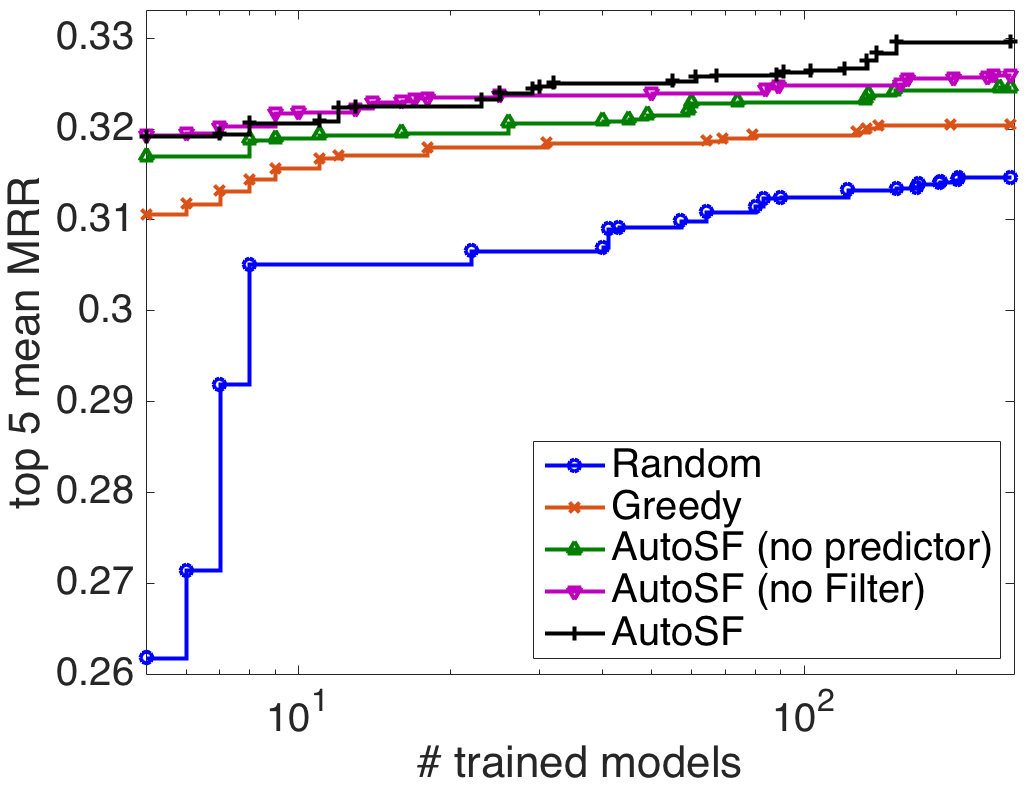}
	\vspace{-10px}
	\caption{Ablation study on the predictor and filter in AutoSF.}
	\label{fig:filpre}
	\vspace{-8px}
\end{figure}

To show the effectiveness of the filter and predictor, 
we remove them from AutoSF and make comparisons in 
%\footnote{+++ the subcaption is removed.}
Fig.~\ref{fig:filpre}.
As shown,
the greedy algorithm is more efficient than random search.
Both filter and predictor are important.
Removing either the filter or predictor will lead to degenerated efficiency.
Besides,
compared with \textit{Greedy},
i.e. no filter and no predictor,
they can both improve efficiency through reducing the search space.

\subsubsection{SRF features} 
\begin{figure}[ht]
	\centering
	\vspace{-8px}
	\includegraphics[height=3cm]{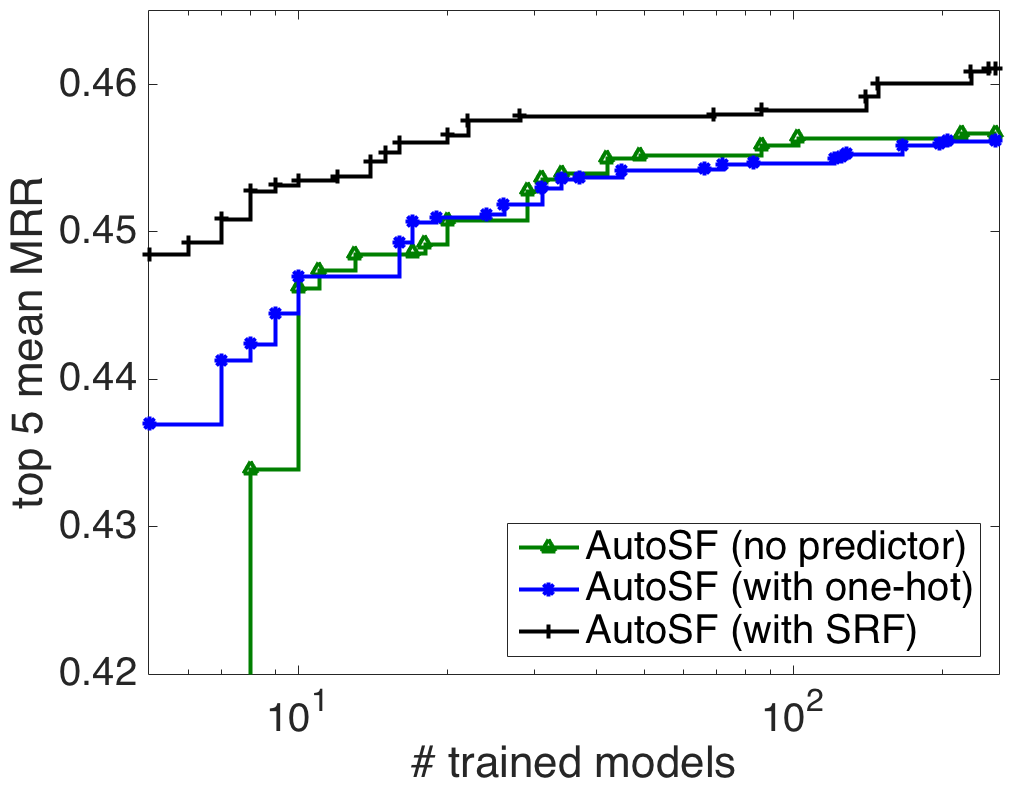}
	\quad
	\includegraphics[height=3cm]{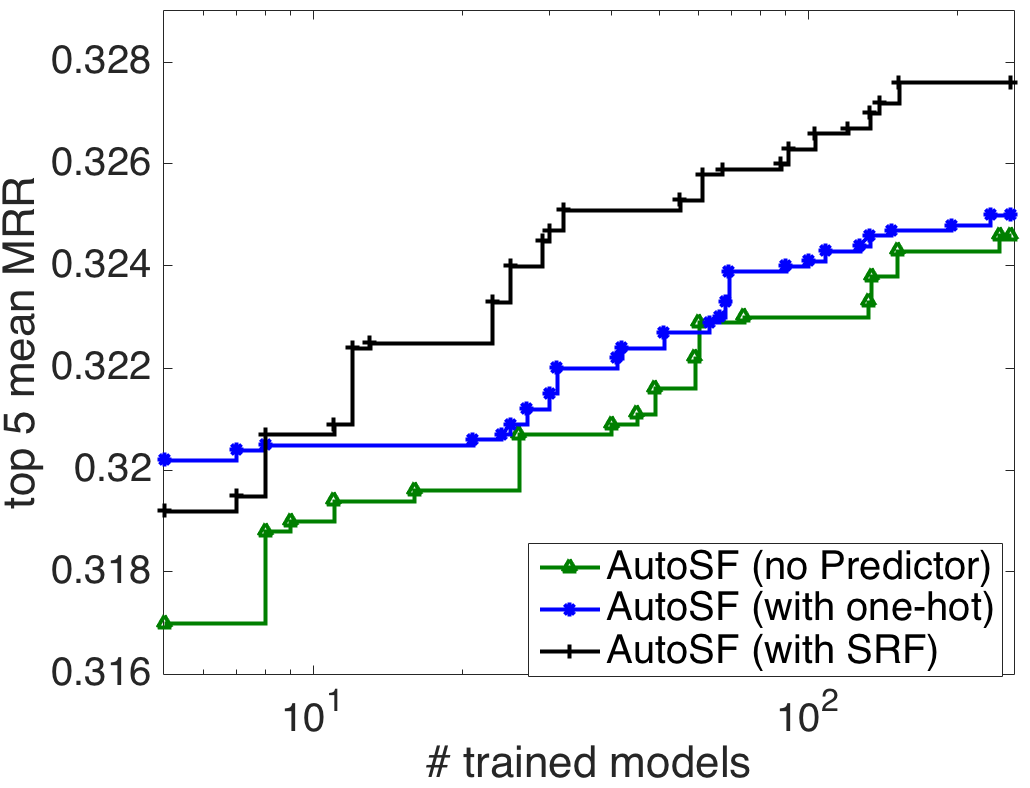}
	\vspace{-10px}
	\caption{Comparison of the proposed SRF with one-hot encoding in \cite{liu2018progressive}.}
	\label{fig:srf}
	\vspace{-5px}
\end{figure}

As in Sec.\ref{ssec:comprand},
one-hot representation \cite{liu2018progressive} can also be used as an alternative to SRFs.
We compare the two kind of features in Fig.~\ref{fig:srf}.
For AutoSF (with one-hot), a 96-8-1 fully connected neural network is used,
and a 22-2-1 network is used for AutoSF (with SRF).
AutoSF (no predictor) is shown here as a baseline 
and it is the same as that in Fig.~\ref{fig:filpre}.
%To further show the effectiveness of SRF compared with one-hot features used in \cite{liu2018progressive},
%we compare the two kind of feature
%we define two metrics here:
%\footnote{$\surd$ I think Fig.~\ref{fig:srf} is enough here.
%	no need to show so many measurements.}

%As for a good predictor, the relative rank it gives for an unseen SF is the main thing we care about
%rather than the prediction value.
%We mainly compare the top-$K_2$ Mrr 
%(note that the Mrr here is specifically designed for comparison in this part,
%and different from the metric we used to evaluate SFs)
%since how well the predictor can select top performed SFs is more important.
%SRC, which is used in PNAS \cite{liu2018progressive}, is added as a reference metric.
%We firstly use the five unique SFs in $f^4$ as the training samples
%and gradually add 16 $f^6$s each interval into the training set.
%All values are evaluated based on $N$ unseen candidates in $f^6$ through greedy and filter.
%In this part, we use $K_2=8$ and $N=256$, same as the experiment setting.
%As shown in Tab.~\ref{tab:srf}, SRF performs generally better than one-hot features,
%especially on WN18RR.
%The one-hot representation not only has higher complexity,
%but also lacks the domain specific properties, namely symmetric or anti-symmetric.
%One-hot feature performs relatively better on FB15k237, whose symmetric property is not as obvious as the others,
%than WN18RR.
%Combining SRF with one-hot feature through data augmentation is an improvement direction
%and we leave it as a future work.

\subsubsection{Sensitivity of meta hyper-parameters}
\label{ssec:hyper}

\begin{figure}[ht]
	\centering
	\includegraphics[height=3cm]{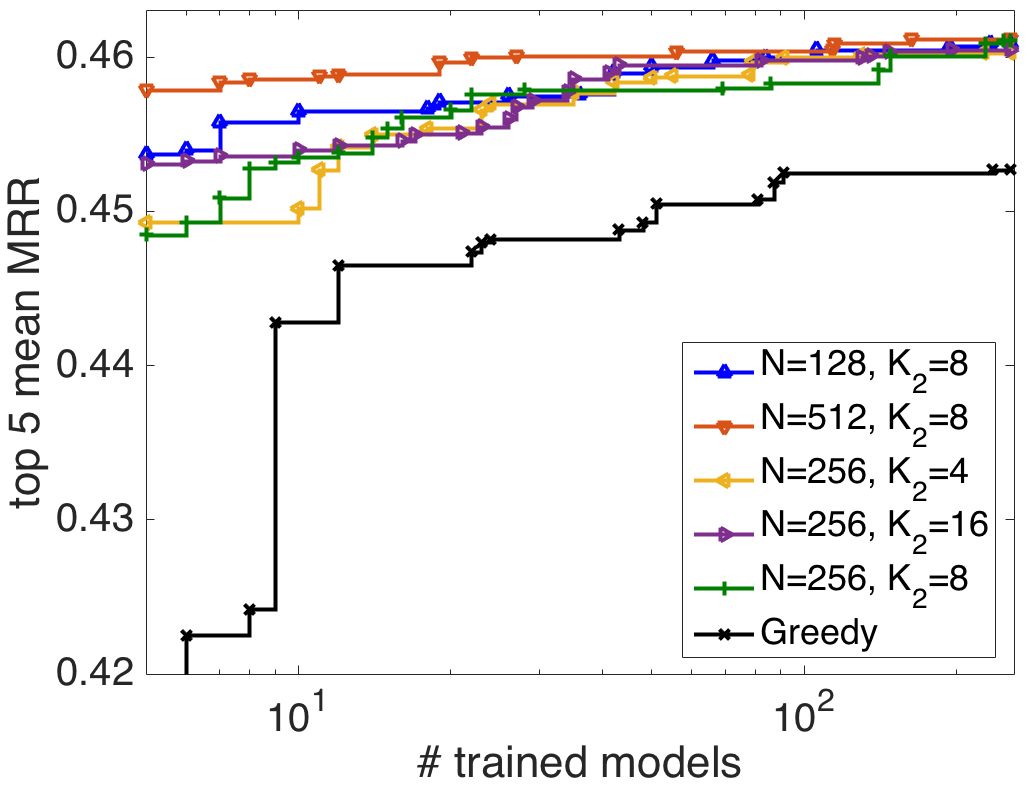}
	\quad
	\includegraphics[height=3cm]{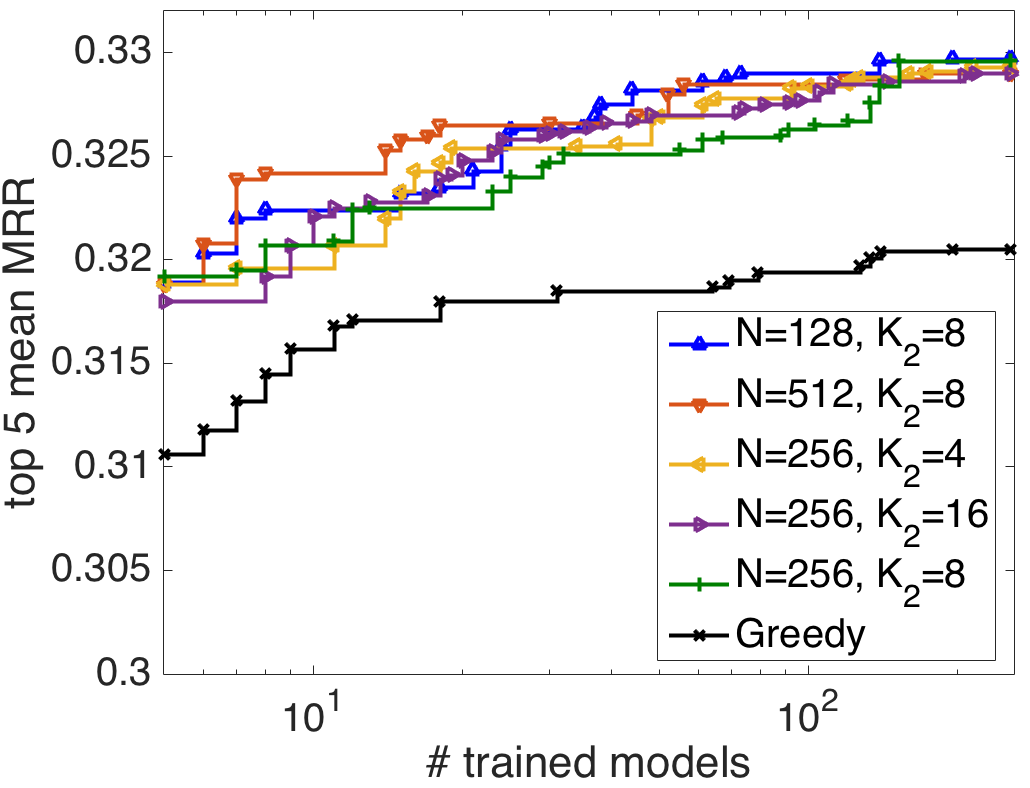}
	\vspace{-10px}
	\caption{Comparison of different meta hyper-parameters. 
		Greedy is added here as a contrast. 
		256 models are evaluated in each setting.}
	\label{fig:hyper}
	\vspace{-8px}
\end{figure}

%\footnote{+++ meta hyper-parameter}
There are three meta hyper-parameters $N, K_1, K_2$ used in our search Alg.\ref{alg:greedy}.
The results we reported in previous parts are based on $N=256, K_1=8, K_2=8$.
We change the value of $N$ to 128 and 512,
$K_2$ to 4 and 16, and show the searching curve on $f^6$ in Fig.~\ref{fig:hyper}.
The parameter
$K_1$ 
that selects top candidates in $f^{b-2}$
%in step~3 of Alg.\ref{alg:greedy}
is not compared for $b=6$ since there are only 5 candidates in $f^4$.
As can be seen,
all the different settings perform similar and obviously outperform the Greedy baseline.

{
\subsubsection{Running time analysis}
\label{ssec:time}

%\footnote{+++ see new Table~\ref{tab:new}.
%	I think it is better organized in this way.}
We show the running time of different components in AutoSF in 
Tab.~\ref{tab:efficiency}.
First, the filter and the predictor (including SRF computation) take
much shorter running time compared with that of the model training.
Then, 
as each greedy step contains 256 model training,
the best SFs can be searched within only several hours (on 8 GPUs), 
except for YAGO3-10
which takes more than one day to evaluate 128 candidates.
%However, the best SF searched in $f^6$ already outperforms the human-designed SFs on YAGO3-10.
%and each greedy step contains 256 model training.
In comparison, 
search problem based on 
reinforcement learning \cite{zoph2017neural} runs over 4 days across 500 GPUs;
genetic programming \cite{xie2017genetic} takes 17 days on single GPU;
and Bayes optimization \cite{feurer2015efficient} trains for several days on CPUs.
Thus,
the proposed AutoSF makes the search problem on KGE tractable,
and it is very efficient in 
AutoML literature.
}

\begin{table}[ht]	
	\centering
	\vspace{-10px}
%	\color{blue}
	\caption{RUNNING TIME ANALYSIS. 
		We show the running time (min) per greedy step (step~2-11 in Alg.\ref{alg:greedy}).
		Apart from step~2-6 (filter), step~7,10-11 (predictor), step~8 (train)
		and step~9 (evaluation),
		all other steps take less than 0.1 minutes.}
	\vspace{-5px}
	\label{tab:stepcost}
	\renewcommand{\arraystretch}{1.1}
	\begin{tabular}{c | c | c | c | c}
		\hline
		\multirow{2}{*}{steps} & filtering & predictor & train           & evaluate \\ \cline{2-5}
		& 2-6       & 7,10-11   & 8               & 9        \\ \hline
		WN18          &  15.9$\pm$0.5   &   1.8$\pm$0.1  & 475.9$\pm$9.5  &  41.3$\pm$0.8    \\ \hline
		FB15K          &   16.8$\pm$0.7  &   1.9$\pm$0.1 & 886.3$\pm$21.8 &  153.7$\pm$3.9   \\ \hline
		WN18RR         &   16.1$\pm$1.0  &   1.8$\pm$0.1   & 271.4$\pm$5.1   & 27.9$\pm$0.5         \\ \hline
		FB15k237        &    16.6$\pm$1.1 &  1.9$\pm$0.1   & 439.2$\pm$11.2  &  63.5$\pm$1.9     \\ \hline
		YAGO3-10        &  16.6$\pm$0.9    &  1.7$\pm$0.1   & 1631.1$\pm$85.5 &  141.9$\pm$8.9        \\ \hline
	\end{tabular}
%	\begin{tabular}{C{1.2cm}|C{0.75cm}|C{0.75cm}|C{1.1cm}|C{1.15cm}|C{1.3cm}}
%		\hline
%		       data sets        &       WN18        &       FB15k        &      WN18RR      &     FB15k237      &     YAGO3-10      \\ \hline\hline
%		\multirow{2}{*}{step~6} & $517.3$ $\pm10.3$ & $1042.7$ $\pm25.7$ & $299.3$ $\pm5.6$ & $504.8$ $\pm13.1$ & $1773$ $\pm 94.5$ \\ \hline
%		       step~2-5         &                                 \multicolumn{5}{c}{16.6$\pm$1.1}                                  \\ \hline
%		        step~8          &                                  \multicolumn{5}{c}{1.8$\pm$0.1}                                  \\ \hline\hline
%	\end{tabular}
	\label{tab:efficiency}
	\vspace{-3px}
\end{table}

In addition,
since AutoSF search SFs with dimension $64$ and then fine-tune the hyper-parameters with $d\in\{256$, $512$, $1024$, $2048\}$ as in \cite{kazemi2018simple,lacroix2018canonical}.
In comparison, the searching cost is comparable with the fine-tune cost
which generally needs to train and evaluate hundreds of hyper-parameter settings with large dimension size.
In this view, the searching cost is not that expensive.

\section{Conclusion}
%\vspace{-10px}
\label{sec:conclusion}
In this paper, we propose AutoSF, an algorithm to automatically design and discover better SFs for KGE.
By using a progressive greedy search algorithm enhanced by a filter and a predictor with domain-specific knowledge,
AutoSF can efficiently design promising SFs
that are KG dependent, new to the literature,
and outperform the state-of-the-art SFs designed by humans
from the huge search space.
{In future work,
a promising direction is to explore how to efficiently search the network structure for NNMs 
under domain-specific constraints.}
The greedy algorithm used in AutoSF somehow limits the exploration in the search space,
which is also a potential problem to be addressed.
%Besides,
%fast evaluation methods like early stopping can be combined with the proposed one 
%to further improve efficiency.
%Besides,
%it will also be interesting to search for SFs that are relation-dependent.
%\footnote{+++ greedy algorithm can explore limited search space,
%	mention that this is a potential problem of the propose method,
%	and we will address this issue in the future.}
%\footnote{+++ also we can mention
%	other fast evaluation methods,
%	i.e.  ``Early stopping methods can help improve efficiency. This is not considered in this paper''.
%	Such methods can also be combine with the propose one
%	for further sped up.}
%\footnote{+++ you can check my old paper to see how to place
%	and ref to appendix.
%	\url{https://arxiv.org/pdf/1710.07205.pdf}}

%\clearpage

\section*{Acknowledgment}
%The work is partially supported by the Hong Kong RGC GRF Project  16202218, 
%the National Science Foundation of China (NSFC) under Grant No. 61729201, 
%Science and Technology Planning Project of Guangdong Province, China, No. 2015B010110006,  
%Hong Kong ITC ITF grants  ITS/391/15FX  and  ITS/212/16FP,  
%Didi-HKUST joint research lab project, 
%Microsoft Research Asia Collaborative Research Grant and 
%Wechat Research Grant.

The work is partially supported by 
the Hong Kong RGC GRF Project  16202218, 
CRF project  C6030-18G,  
AOE project AoE/E-603/18,  
the National Science Foundation of China (NSFC) under Grant No. 61729201, 
Science and Technology Planning Project of Guangdong Province, China, No. 2015B010110006,  
Hong Kong ITC grants  ITS/044/18FX  and  ITS/470/18FX,  
Didi-HKUST joint research lab Grant, 
Microsoft Research Asia Collaborative Research Grant, 
Wechat Research Grant and Webank Research Grant.

\bibliographystyle{plain}
\bibliography{bib}

\appendix

%\section{Proofs}
\subsection{Proof of Proposition \ref{pr:expre}}
\label{app:proof:expre}
\begin{proof}
	We consider the four cases separately:
	\begin{itemize}[leftmargin=10px]
		\item \textit{symmetric relations}: 
		If $g(\bm{r})^\top=g(\bm{r})$ for some $\bm{r}\in\mathbb R^d$,
		then given a triplet $(h, r, t)$,
		$	f(\bm{t},\bm{r},\bm{h}) = \bm{t}^\top g(\bm{r})\bm{h} 
		= \left(\bm{t}^\top g(\bm{r})\bm{h}\right)^\top 
		= \bm{h}^\top g(\bm{r})^\top\bm{t} 
		= \bm{h}^\top g(\bm{r})\bm{t} 
		= f(\bm{h}, \bm{r}, \bm{t})$,
		which means $g(\bm{r})$ can handle symmetric relations.
		
		\item \textit{anti-symmetric relations}:
		If $g(\bm{r}')^\top=-g(\bm{r}')$ for some $\bm{r}'\in\mathbb R^d$,
		then given a triplet $(h, r', t)$,
		$f(\bm{t},\bm{r}',\bm{h}) = \bm{t}^\top g(\bm{r}')\bm{h} 
		= \left(\bm{t}^\top g(\bm{r}')\bm{h}\right)^\top 
		= \bm{h}^\top g(\bm{r}')^\top\bm{t}
		= -\bm{h}^\top g(\bm{r}')\bm{t} 
		= -f(\bm{h},\bm{r}',\bm{t})$,
		which means $g(\bm{r}')$ can handle anti-symmetric relations.

		\item \textit{general asymmetric relation}:
		Since $\mathbf D_i^{\bm{r}} = \diag{\bm{r}_i}, i=1\dots4$, 
		then for any scalar $w\in R$, 
		$\mathbf D_i^{(w\bm{r})} = \diag{w\bm{r}_i} = w\mathbf D_i^{\bm{r}}$.
		This leads to $g(w\bm{r}) = wg(\bm{r})$.
		Similarly, we have $g(w_1\bm{r} + w_2\bm{r}') = w_1g(\bm{r}) + w_2g(\bm{r}')$ for scalar $w_1, w_2$.
		
		If $g(\bm{r})^\top=g(\bm{r})$ for some $\bm{r}\in\mathbb R^d$ and $g(\bm{r}')^\top=-g(\bm{r}')$ for another $\bm{r}'\in\mathbb R^d$, then for any general assymetric relation $r^{\text{asym}}$, let $\bm{r}^{\text{asym}} = w_1\bm{r} + w_2\bm{r}'$. We have
		\begin{align}
		& f(\bm{h},\bm{r}^{\text{asym}},\bm{t}) 
		= \bm{h}^\top g(\bm{r}^{\text{asym}})\bm{t}  
		= \bm{h}^\top g(w_1\bm{r} + w_2\bm{r}')\bm{t}  
		\label{eq:asym:forward}\\
		&= w_1\bm{h}^\top \!\! g(\bm{r})\bm{t} 
		\! + \! w_2 \bm{h}^\top \!\! g(\bm{r}') \bm{t}  
		\! = \! w_1 f(\bm{h}, \bm{R}, \bm{t}) \! + \! w_2 f(\bm{h},\bm{r}',\bm{t}). 
		\notag
		\end{align}
		Similarly, we can obtain
		\begin{align}
		f(\bm{t},\bm{r}^{\text{asym}}, \bm{h}) 
		&= w_1 f(\bm{t},\bm{r},\bm{h}) + w_2 f(\bm{t}',\bm{r}',\bm{h}) \nonumber  \\
		&= w_1 f(\bm{h}, \bm{r}, \bm{t}) - w_2(\bm{h}, \bm{r}', \bm{t}).
		\label{eq:asym:backward}
		\end{align}
		Then, for any value of the pair $f(\bm{h},\bm{r}^{\text{asym}},\bm{t})$ 
		and $f(\bm{t},\bm{r}^{\text{asym}}, \bm{h})$, 
		there exist appropriate scalars $w_1, w_2$ by solving \eqref{eq:asym:forward} and \eqref{eq:asym:backward} 
		to obtain $\bm{r}^{\text{asym}} = w_1\bm{r} + w_2\bm{r}'$.
		
		\item \textit{inverse relations}:
		Let $r^{a}$ and $r^{b}$ be two relations, 
		and assume $\bm{r}^{a} = w_1\bm{r} + w_2\bm{r}'$ 
		and $\bm{r}^b = w_3\bm{r} + w_4 \bm{r}'$
		given $g(\bm{r})^\top =g(\bm{r}), g(\bm{r}')^\top =-g(\bm{r}')$ based on general asymmetric property.
		Let $w_1=w_3$ and $w_2=-w_4$, then
		\begin{align*}
		%\vspace{-5px}
		\!\!\!\!\!\!
		f(\bm{t}, \bm{r}^a, \bm{h}) 
		&= w_1 f(\bm{t},\bm{r},\bm{h}) + w_2f(\bm{t},\bm{r}',\bm{h}) \\
		&= w_1 f(\bm{h}, \bm{r}, \bm{t}) - w_2 f(\bm{h},\bm{r}',\bm{t}) \\ 
		&= w_3 f(\bm{h}, \bm{r}, \bm{t}) + w_4 f(\bm{h},\bm{r}',\bm{t}) = f(\bm{h}, \bm{r}^b, \bm{t}),
		\\
		g(\bm{r}^a) &= w_1 g(\bm{r}) + w_2 g(\bm{r}')  
		= w_1 g(\bm{r})^\top - w_2 g(\bm{r}')^\top \\
		&= w_3 g(\bm{r})^\top \!\!\! + \! w_4 g(\bm{r}')^\top \!\!
		\! = \! g(w_3\bm{r} \! + \! w_4\bm{r}')^\top
		\!\!\! = \! g(\bm{r}^b)^\top \!
		\!\!.
		\vspace{-4px}
		\end{align*}
		This means $r^a$ and $r^b$ are a pair of inverse relations.
	\end{itemize}
	Thus,
	we obtain the proposition.
\end{proof}

\subsection{Proof of Proposition \ref{pr:srfs}}
\label{app:proof:srfs}
\begin{proof}
	(i) For each case (S1-11), the SRF is generated based on permutation and flipping signs of the 4 basic values $[v_1; v_2; v_3; v_4]$.
	Thus, no matter how the structure $g$ changes due to permutation or flipping signs,
	they will lead to the same feature.
	Once the matrix $g(\mathbf v)$ can be symmetric or anti-symmetric under one assignment S$i$,
	its corresponding feature will not change regardless of permutation or flipping signs.
	(ii) 
	The SRF is generated based on the symmetric and skew-symmetric property
	and each dimension corresponds to a specific case of symmetric or skew-symmetric.
	Then the predictor can learn higher weights to the dimensions correlates the data's symmetric property well.
	Besides, this pattern can be easily learned through  a few samples.
\end{proof}

\subsection{Design of SRFs}
\label{app:srf}
%and are also invariant to permutation/flipping signs.
%The SRFs we designed are inspired by following key observation on $g(\bm{R})$.
%Let us look at
%the example in Fig.~\ref{fig:invariance}(a) again,
%%\footnote{+++ need to be consistent, check until}
%if $\mathbf{D}^{\bm{R}}_3 = \mathbf{D}^{\bm{R}}_2, \mathbf{D}^{\bm{R}}_4 = \mathbf{D}^{\bm{R}}_1$ with proper $\bm{r}$,
%then $g(\bm{R}) - g(\bm{R})^{\top} = \mathbf 0$,
%which means such $g(\bm{R})$ can be symmetric;
%if $\mathbf{D}^{\mathbf{r'}}_3 = -\mathbf{D}^{\mathbf{r'}}_2$ 
%and $\mathbf{D}^{\mathbf{r'}}_4 = -\mathbf{D}^{\mathbf{r'}}_1$ with proper $\bm{r}'$,
%then $g(\bm{R}') + g(\mathbf{r'})^{\top} = \mathbf 0$,
%which means such $g(\bm{R}')$ can be skew-symmetric.
%%Based on Proposition~\ref{pr:expre}, the SF defined by Fig.~\ref{fig:invariance}(a) is a good candidate.
%However, how to obtain $\bm{r}$ with appropriate values
%to check the symmetric property of $g$ is still a problem
%since the values are not known in advance.
%Fortunately, the general structure of $g$ can be abstracted as a small $4\times 4$ matrix
%when $\mathbf D^{\bm{r}}_i$ reduces to be 1-dimensional.
%Thus, we can directly assign proper scalar value $v_i$ for each $\mathbf D^{\bm{r}}_i$ as in Fig.~\ref{fig:feat-gen}.
%Then let $\mathbf v = [v_1; v_2; v_3; v_4]$,
%the symmetric and skew-symmetric property of $g$
%can be efficiently checked through $g(\mathbf {v}) \! - \! g(\mathbf{v})^{\top}$ 
%and
%$g(\mathbf {v}) \! + \! g(\mathbf {v})^{\top}$ since $g(\mathbf v)$ here is a simple $4\times 4$ matrix.

\begin{remark}[SRF] \label{def:SRFs}
	%	%\vspace{-4pt}
	Let the 1-dimensional degeneration of $\mathbf D_1^{\bm{r}}$, $\mathbf D_2^{\bm{r}}$, $\mathbf D_3^{\bm{r}}$, $\mathbf D_4^{\bm{r}}$ be scalars $v_1, v_2, v_3, v_4$.
	%	Let $\bm{R}_1$ to $\bm{R}_4$ be scalars as $r_1$ to $r_4$.
	We give $\mathbf v=[v_1; v_2; v_3; v_4]$ with following assignments:
	\begin{itemize}[leftmargin=10pt,topsep=0pt,parsep=0pt,partopsep=0pt]
	\item Four values are non-zero:
	Four values are non-zero: (S1). All of them have different absolute value, like $[1;2;3;4]$;
	(S2). Two have the same absolute value, and the other two have another same absolute value,	like $[1;1;2;2]$;
	(S3). Two of them have the same absolute value while the other two not, like $[1;1;2;3]$;
	(S4). Three of them have the same absolute value while another one not, like $[1;1;1;2]$; 
	(S5). All have the same absolute value, like $[1;1;1;1]$.
	
	\item Three values are non-zero:
    (S6). All of them have different absolute value, like $[0;1;2;3]$;
    (S7). Two of them have same absolute value, like $[0;1;1;2]$;
    (S8). All of them have same absolute value, like $[0;1;1;1]$.
    
    \item Two values are non-zero:
    (S9). They have different absolute value, like $[0;0;1;2]$;
    (S10). They have the same absolute value, like $[0;0;1;1]$.
    
    \item Only one is non-zero: (S11). $[0;0;0;1]$.
	\end{itemize}
	\noindent
	For each (S1)-(S11),
	we use permutation and flipping the signs based on the given examples to check if $g(\mathbf v)$ can be symmetric or skew-symmetric under each case.
	As a result, a $11\times2 = 22$ dimensional SRF returns with little extra cost.
\end{remark}

The 11 cases exhaustively enumerate the possible conditions of $g(\cdot)$ being symmetric or skew-symmetric.
What we care most is what kind of symmetric properties $g(\bm{r})$ can be under these cases.
Some data sets may need more features being 1 if it has more symmetric, anti-symmetric and inverse relations like FB15k,
but some may not like FB15k237.
The process of SRF generation is given in Alg.\ref{alg:srf}.

\begin{algorithm}[ht]
\caption{SRF generation for each (S1-11)}
\label{alg:srf}
\small
\begin{algorithmic}[1]
	\REQUIRE the structure of $g$, $\text{SRF}_i$=[0,0] for $i=1\dots11$; 
	\FOR{$\mathbf v$ in the assignment candidates of S$i$ through permuting and flipping signs}
	\IF {$g(\mathbf v) -  g(\mathbf v)^\top = \mathbf 0$}
	\STATE $\text{SRF}_i$[0] = 1 \quad // symmetric
	\ENDIF
	\IF{$g(\mathbf v) + g (\mathbf v)^\top = \mathbf 0$}
	\STATE $\text{SRF}_i$[1] = 1 	\quad // skew-symmetric
	\ENDIF
	\ENDFOR
	\RETURN $\text{SRF}_i$ \qquad // 2-dimensional components for each S$i$.
\end{algorithmic}
\end{algorithm}

\section{Additional Materials for the Experiments}

\subsection{Details of Taking MLP as $\mathcal{G}$}
\label{app:mlp}

To ensure quick training and testing \cite{dettmers2017convolutional},
we use two fully-connected neural networks as the MLP.
Specifically, to predict the tail entity, we use $NN_1$ to combine $\bm{h}$ and $\bm{r}$ into $\mathbf v = NN_1(\bm{h}, \bm{r})$.
Then we use the dot product of $\mathbf v$ and $\bm{t}$ as the score.
To test the head entity, another network $NN_2$ is built in similar way and final score is 
$\left\langle NN_2(\bm{t}, \bm{r}), \bm{h}\right\rangle$.
The two networks share the same structure (128-64-64) and are trained jointly based on Alg.\ref{alg:general}.

\end{document}